\ifdefined\XeTeXversion\else\pdfoutput=1\fi 
\documentclass{article} %
\usepackage[T1]{fontenc} %
\usepackage{iclr2027_conference,times}

\usepackage{amsmath,amsfonts,bm}

\def\eqref#1{equation~\ref{#1}}

\def\1{\bm{1}}

\DeclareMathAlphabet{\mathsfit}{\encodingdefault}{\sfdefault}{m}{sl}
\SetMathAlphabet{\mathsfit}{bold}{\encodingdefault}{\sfdefault}{bx}{n}

\usepackage{hyperref}
\usepackage{url}
\usepackage{xcolor}
\usepackage{colortbl}
\usepackage{graphicx}
\usepackage{booktabs}
\usepackage{multirow}
\usepackage{float}
\usepackage{placeins}

\newcommand{\first}[1]{\cellcolor{red!25}\textbf{#1}}
\newcommand{\second}[1]{\cellcolor{orange!25}\textbf{#1}}
\newcommand{\third}[1]{\cellcolor{yellow!35}\textbf{#1}}

\title{Bake It Till You Make It: Ultrafast\\Spatial texture atlas Splatting}

\author{Neel Kelkar$^{1,2}$ \quad Simon Niedermayr$^{1}$ \quad Kaloian Petkov$^{2}$ \\
\textbf{Klaus Engel}$^{2}$ \quad \textbf{R\"udiger Westermann}$^{1}$ \\
$^{1}$Technical University of Munich \quad $^{2}$Siemens Healthineers \\
\texttt{\{neel.kelkar, simon.niedermayr, westermann\}@tum.de} \\
\texttt{\{kaloian.petkov, engel.klaus\}@siemens-healthineers.com} \\
Project page: \url{https://nilkel.github.io/bitymi/}
}

\iclrfinalcopy
\renewcommand{\iclrheadertext}{Preprint.}

\begin{document}

\maketitle

\begin{abstract}

Neural radiance representations in Gaussian Splatting (GS) deliver high-fidelity color detail but impose substantial rendering overhead from network evaluations. We present a learned sparse scene representation based on 2D surfels, which enables baking the neural component into a hardware-accelerated 2D texture atlas and eliminates runtime network inference. 
Our 2D surfels carry low-frequency geometry and view-dependent appearance, while view-independent, per-primitive high-frequency texture is encoded with a spatial hash grid and converted into the texture atlas. A novel sparsity objective that penalizes per-primitive kernel falloffs achieves a substantially sparser representation than prior methods. With a compute-optimized ray-surfel intersection shader our approach renders roughly $7$ to $9\times$ faster than 3DGS on common benchmarks, and over an order of magnitude faster on individual scenes, while surpassing the perceptual quality and speed of the fastest sparsification methods (FastGS and Speedy-Splat) at PSNR parity. A quality-optimized variant matches the perceptual quality of the strongest baseline while still rendering several times faster. Because inference is one texture fetch per fragment, frame rates can be pushed to $2{,}700$--$4{,}400$ FPS with no loss in quality when using the GPU hardware rasterizer.

\end{abstract}

\begin{figure}[h]
  \centering
  \includegraphics[width=\textwidth]{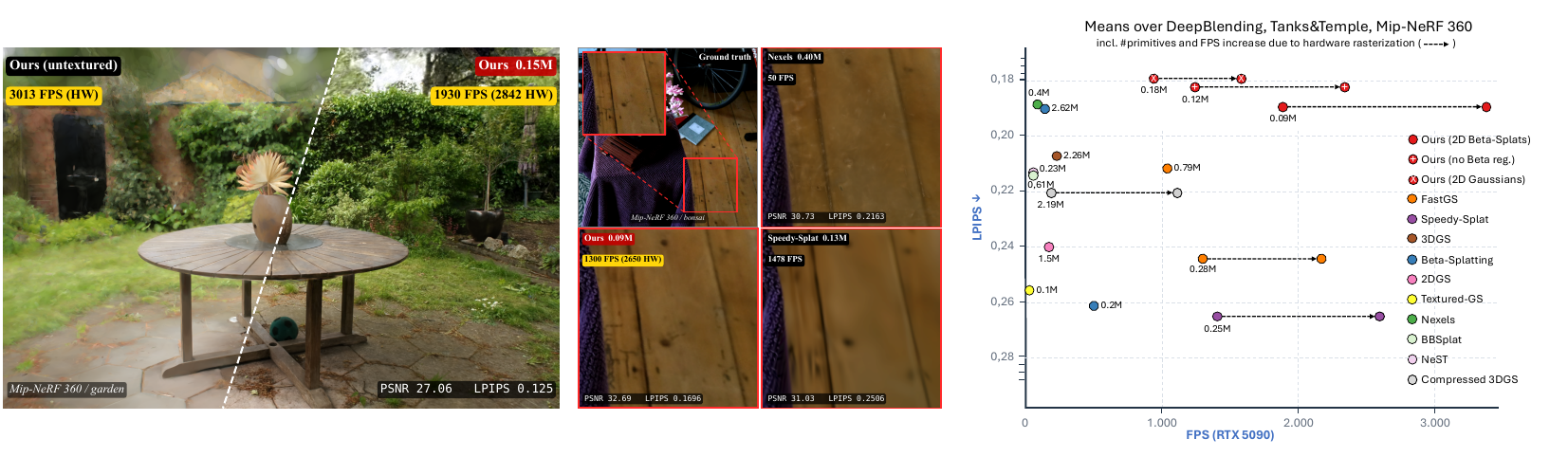}
  \vspace{-20pt}
  \caption{\emph{Left:} Ours wo/w learned view-independent appearance features. \emph{Middle:} Ours compared against the highest quality and fastest baselines. 
  \emph{Right:} Quality--speed landscape of GS variants; the comparison videos on the project page show temporal consistency comparable to prior work using hardware rasterization.
  }
  \label{fig:teaser}
\end{figure}

\section{Introduction}
3D Gaussian Splatting (3DGS) \citep{kerbl20233d} represents a scene as an explicit, unstructured set of anisotropic Gaussians that are rasterized and alpha-blended directly onto the image plane, and it has become one of the strongest tools for novel view synthesis (NVS). Its persistent limitation is texture. Each Gaussian contributes a spatially smooth basis function, so high-frequency appearance forces the optimization to place many small Gaussians densely across textured surfaces. Grid-based methods store such detail in voxel arrays or hash-encoded latent vectors \citep{mildenhall2021nerf, muller2022instant}; 3DGS has no comparable spatially local feature encoding, and its primitive count is tied to the frequency content of the scene's appearance.

Textured Splatting approaches disentangle appearance from geometry with two families of methods. Explicit textured methods like BBSplat \citep{svitov2025billboard}, Textured Gaussians \citep{chao2025textured}, Texture-GS \citep{xu2024texture}, GSTex \citep{rong2025gstex}, and Content-Aware Texturing \citep{papantonakis2025content} all explicitly optimize per-primitive texture representations but are bound in memory by the primitive count and try to learn 3D textures with a 2D representation.  Neural methods such as NeST Splatting \citep{zhang2025neural}, Nexels \citep{rong2025nexels}, and Hybrid Latents~\cite{kelkar2026hybrid} query a global multi-resolution hashgrid at the ray--surfel intersection for appearance.
The hash-grid provides stable convergence and caps the texture's memory overhead during training. However, since every fragment incurs a neural field query, rendering is significantly slowed down.

We combine 2D surfels with a radiance formulation that separates low-frequency geometry and view-dependent appearance, carried by the surfels, from view-independent appearance, carried by a hash grid as an unbounded residual. Because the residual is view-independent, it can be baked into a single explicit texture atlas of one rectangle per surfel. Using negative and positive color offsets for the residuals provides more representational power. The texels are then fine-tuned against the training images to overcome hash collisions from training. 

A texture atlas alone does not make rendering fast, since frame time is bounded by sorting and blending the textured Gaussians and performance quickly suffers from overdraw. We therefore train the field jointly with the multi-view error criterion of FastGS \citep{ren2025fastgs} to limit densification. Furthermore, we introduce a falloff prior using Beta kernels \citep{liu2025deformable}, which progressively regularizes the kernels toward flat, hard-edged profiles. Flat kernels stop redundant overlaps between soft Gaussian tails and dampen positional gradients, reducing both primitive count and overdraw.

With our sparse surfel representation we accept 55--320\,MB texture atlases for typical scenes to gain sparsity and speed. These are larger than memory-optimized 3DGS scene representations~\citep{niedermayr2024compressed}, but run without issues even on low-to-mid-range GPUs. Using hardware texture mapping gives $5.4$--$6.9\times$ the frame rate of 3DGS with the standard 2DGS ray--surfel shader~\citep{Huang_2024}. Since at these rates the shader becomes a performance bottleneck, we derive an exact rational reformulation of the intersection test, raising the speedup to $7.1$--$9.1\times$ over 3DGS and $1.2$--$1.8\times$ over FastGS and Speedy-Splat at better perceptual quality, and to $14.5\times$ over Beta-Splatting, the most accurate baseline. Using the GPU's hardware rasterizer further speeds up rendering of $1.8\times$. In summary, our contributions are:
\begin{itemize}
\item A neural view-independent appearance representation for unbounded texture residuals.
\item A regularizer progressively flattening surfel kernels, yielding sparser geometry and less overdraw.
\item A baking procedure from a neural representation to a compact RGB texture atlas, followed by direct finetuning of the texels.
\item A closed-form ray-surfel intersection that removes the per-pixel intersection solve from the software rasterizer.
\end{itemize}

\section{Related Work}
\textbf{Primitives and kernels.} 3DGS~\citep{kerbl20233d} models scenes as unstructured sets of volumetric Gaussians rendered by differentiable splatting. Soft, unbounded Gaussians bloat beyond surface boundaries, so 2DGS~\citep{Huang_2024} flattens them into oriented planar surfels, Triangle Splatting~\citep{held2025triangle} and 3D Convex Splatting~\citep{held20253d} replace them with convex shapes, and Deformable Beta kernels~\citep{liu2025deformable} morph between soft blobs and hard-edged disks. For view-dependent color, Spherical Harmonics are band-limited and can miss sharp highlights. Spherical Gaussians~\citep{wang2026sg} and Spherical Betas~\citep{liu2025deformable} fit specularities with localized lobes; Spherical Voronoi partitions~\citep{di2025spherical} give an explicit parameterization that covers both diffuse shifts and sharp reflections, and score highest in our comparison on \emph{bonsai} (Table~\ref{tab:ablation_spherical}).

\textbf{Sparsification.} Default density control relies on positional gradients and opacity thresholds and leaves many redundant primitives. MiniSplatting~\citep{fang2024mini}, GaussianSpa~\citep{zhang2025gaussianspa}, FastGS~\citep{ren2025fastgs}, and Speedy-Splat~\citep{hanson2025speedy} reduce the count through intersection-preserving simplification, regularization, multi-view error scoring, and tight screen-space bounds with pruning. The surviving primitives must still encode high-frequency texture, which limits the attainable sparsity and the texture quality.

\textbf{Textured primitives.} GStex~\citep{rong2025gstex}, Textured Gaussians~\citep{chao2025textured}, BBSplat~\citep{svitov2025billboard}, and Content-Aware Texturing~\citep{papantonakis2025content} store textures on the primitives themselves. Texture memory then scales with primitive count and with resolutions fixed before training. Because the textures are allocated before densification, their parameters, gradients and optimizer state grow with every clone. Texture-GS~\citep{xu2024texture} maps Gaussian centers to a global unit-sphere UV domain through an MLP, limiting to object-centric scenes.
Pipelines using per-primitive textures start from a converged dense model with densification frozen~\citep{rong2025gstex,chao2025textured}, which caps the attainable sparsity. Alpha textures increase overdraw on top of that.
We allocate the atlas once, after sparsification, so its size follows the primitives that survive rather than limiting training.

The atlas requires about 55--320\,MB for typical scenes, more than memory-optimized 3DGS scene representations \citep{niedermayr2024compressed} need. Extreme compactness, however, comes at the cost of reduced rendering speed and quality, even with hardware rasterization (Sec. \ref{sec:Results}). We compare against Compressed 3DGS in Table~\ref{tab:nvs_results} and Sec.~\ref{sec:hw_baselines}.

\textbf{Neural and Hybrid representations.} NeST-Splatting~\citep{zhang2025neural} offloads appearance to a global hash grid that is queried per fragment with an MLP decoder. Nexels~\citep{rong2025nexels} and Hybrid Latents~\citep{kelkar2026hybrid} combine neural texturing with per-primitive appearance. Due to the compact latent grid these methods require less memory than ours, but keep a slow neural query in the render loop.
Like ours, NeRF-based approaches such as SNeRG~\citep{hedman2021baking}, MERF~\citep{reiser2023merf}, and SMERF~\citep{duckworth2024smerf} bake view-independent neural queries into lightweight 3D features for faster inference, but still require a costly screen-space decoder that limits rendering speed.

\begin{figure}[t!]
  \centering
  \includegraphics[width=1\linewidth]{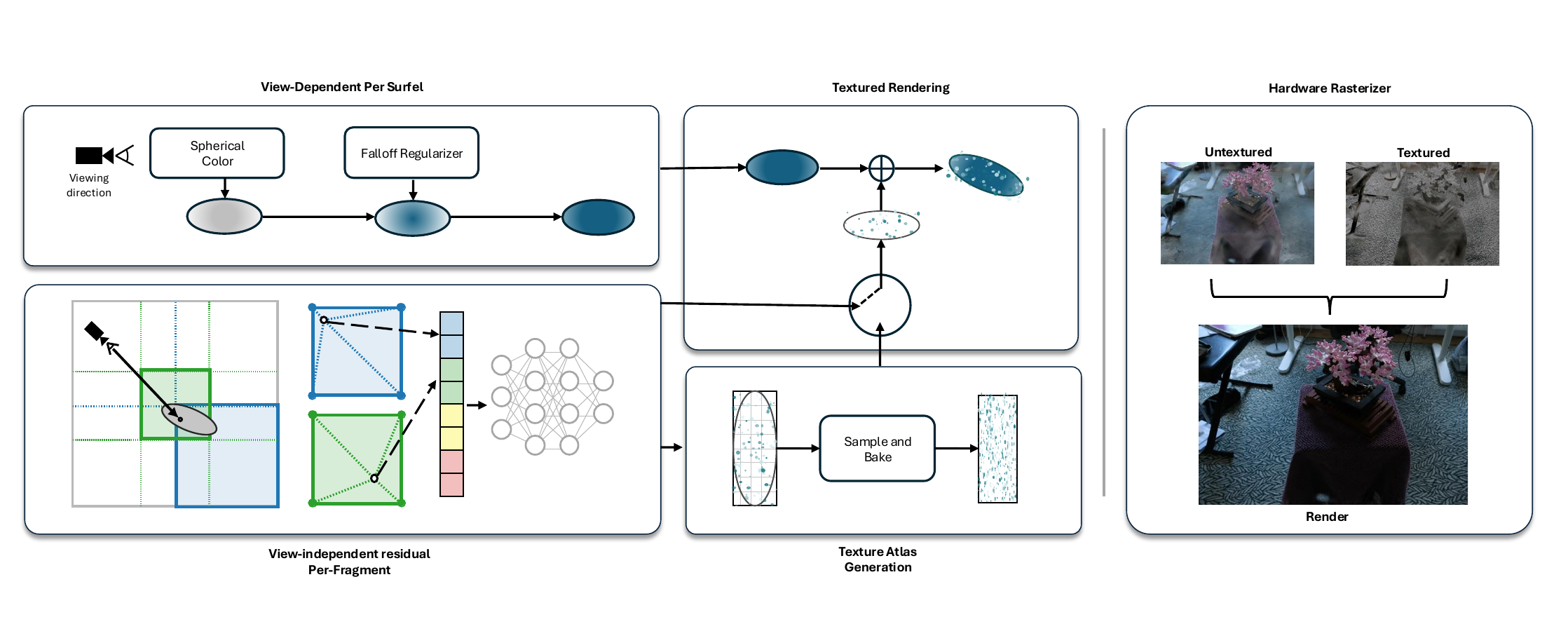}
  \vspace{-30pt}
  \caption{A multi-resolution 3D hash-grid learns view-independent appearance, and a spherical Voronoi model per surfel captures sharp, view-dependent effects. A falloff prior progressively flattens the kernels to reduce overdraw and let the hash grid drive densification. The hash-grid is sampled across surfels, decoded colors are optimized, and baked as residuals in a 2D texture atlas. We render with a hardware rasterizer with accelerated texturing. }
  \label{fig:pipeline}
\end{figure}

\section{Methodology}
We represent a scene as a sparse set of oriented 2D Beta kernels,  regularized to reduce overdraw. 
We separate appearance into per-surfel view-dependent color and a per-fragment texture residual that is learned with a spatial multi-resolution hash-grid and a world-space MLP decoder (see Fig.~\ref{fig:pipeline}).
For inference, we sample the hash-grid across the surface of each primitive, bake the output into a compressed RGB texture atlas, and finetune the atlas texels directly against the training images. Rendering then reduces to fast, per-fragment texture fetches evaluated through an exact rational reformulation of the ray-surfel intersection.

\subsection{Fundamentals}
\label{sec:beta-kernels}
3DGS \citep{kerbl20233d} maps volumetric ellipsoids to the image plane with EWA splatting \citep{zwicker2002ewa}; 2DGS \citep{Huang_2024} flattens them into oriented planar surfels, which we build on. Each surfel $i$ has a position $\mu_i$, rotation $q_i$, two-dimensional scale $s_i$, and opacity $o_i$. A pixel's color is the front-to-back alpha composite of the depth-sorted surfels overlapping it, where surfel $i$ contributes color $c_i(x)$ with effective alpha $\alpha_i(x) = o_i\, G_i(x)$ for kernel $G$ at the ray--surfel intersection $x$, weighted by the transmittance accumulated in front of it. The color $c_i(x)$ may be constant per surfel or vary across its footprint, and the formulation is differentiable in every attribute. We replace the Gaussian kernel of 2DGS with the Deformable Beta kernel \citep{liu2025deformable}
\begin{equation}
\mathcal{B}(x;b) = (1-x)^{\beta(b)},\quad \beta(b)=5\sigma(b),\quad x \in [0,1],\; b \in \mathbb{R}, \label{eq:beta-kernel}
\end{equation}
where $\sigma$ is the sigmoid and $x$ the squared radial distance from the kernel center, normalized to $1$ at the edge of the support. While large $b$ gives soft, Gaussian-like surfels, small $b$ gives flat, hard-edged disks that fit sharp geometry with fewer primitives and less overdraw. We restrict $b$ so that kernels lie between a flat disk and a soft Gaussian.

\subsection{Neural Appearance Decomposition}
\begin{figure}[t]
  \centering
  \begin{minipage}[c]{0.422\linewidth}
    \centering
    \includegraphics[width=\linewidth]{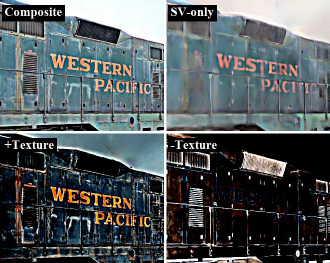}
  \end{minipage}\hfill
  \begin{minipage}[c]{0.558\linewidth}
    \centering
    \includegraphics[width=\linewidth]{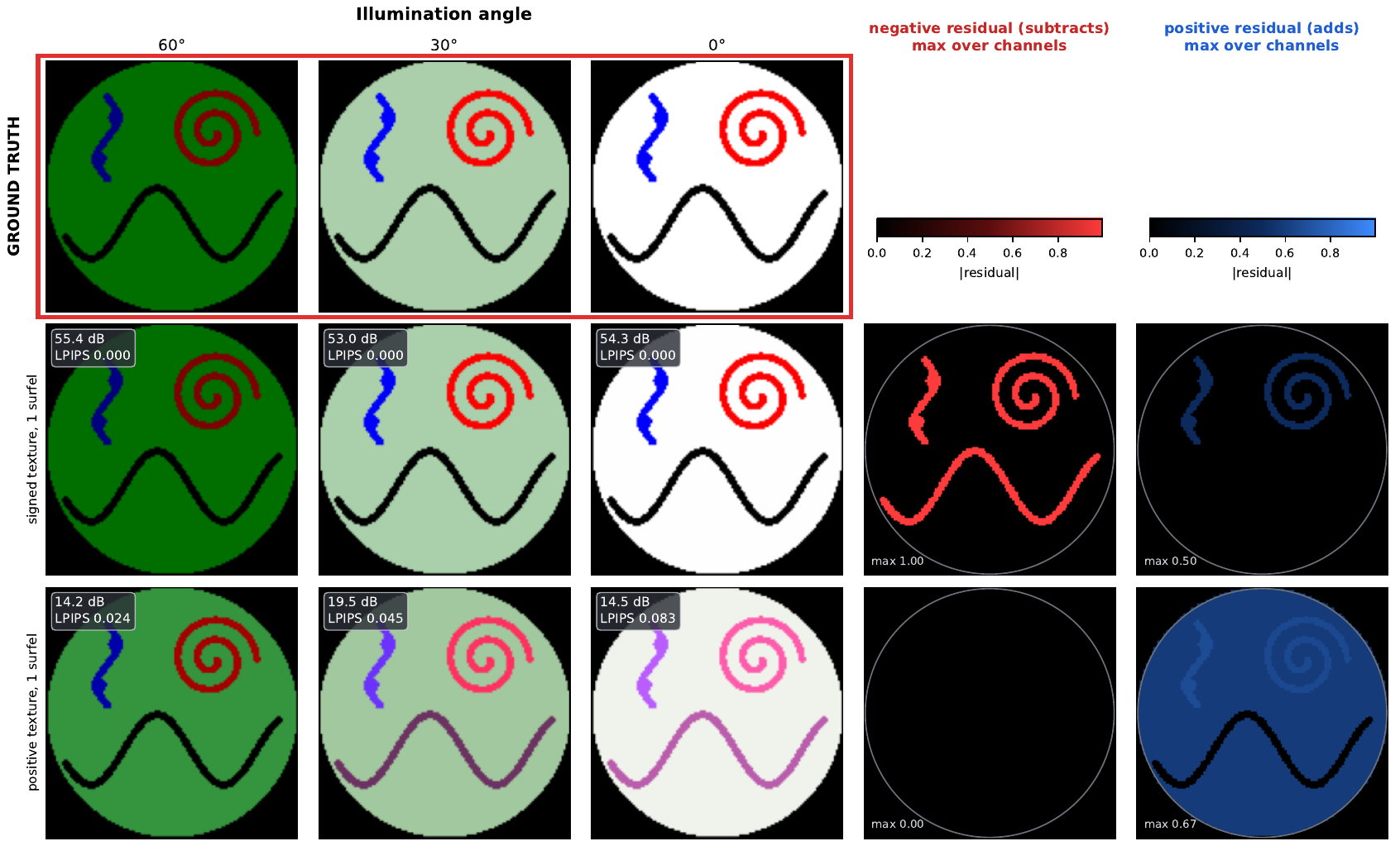}
  \end{minipage}
  \caption{\textbf{Signed texture residual.} \emph{Left:} appearance decomposition on a real scene, in reading order our composited render, the spherical Voronoi term $f_{SV}$ alone, and the positive and negative parts of the residual $\boldsymbol{\rho}$. The residual carries the high-frequency detail the per-surfel term does not represent. \emph{Right:} one textured surfel under three illumination angles. A signed residual reproduces the color variation with a single surfel ($53$--$55$\,dB) where a positive additive texture cannot ($14$--$20$\,dB); the last two columns show the negative and positive parts of each learned residual.}
  \label{fig:decomposition}
\end{figure}

We decompose each surfel's appearance into a per-primitive color $f_{SV}(\mathbf{d})$ depending on viewing direction $d$, and a global view-independent residual $\boldsymbol{\rho}(\mathbf{x}) = f_\phi(E_\theta(\mathbf{x}))$ from a multiresolution hash grid $E_\theta$ and an MLP decoder $f_\phi$. A fixed offset $b_{SV} = 0.5$ inside the inner activation gives a neutral gray baseline on which the residual can build before $f_{SV}$ is trained. The fragment color is
\begin{equation}
\mathbf{c}(\mathbf{x}, \mathbf{d}) = \mathrm{ReLU}\Big(
    \mathrm{ReLU}\big( f_{SV}(\mathbf{d}) + b_{SV} \big)
    + f_\phi\!\big(E_\theta(\mathbf{x})\big)
\Big).
\label{eq:color}
\end{equation}
Gradient descent takes the pathway that lowers the photometric loss fastest. The hash grid is queried at every fragment along the surfel's surface, so it sees sub-surfel coordinate variation and absorbs fine spatial texture; per-surfel attributes are constant across a surfel, so they absorb what does not vary across it. View dependence is low-frequency and shared across a surfel, and it ends up in the per-surfel vector (Fig.~\ref{fig:decomposition}).

\subsubsection{View-independent Surfel Texture}
The residual is unbounded, so the texture can add to or subtract from the base color. This lets one surfel carry detail that a positive-only texture could reproduce only by splitting it (Fig.~\ref{fig:decomposition}; Appendix~\ref{sec:ablation_signed} quantifies the difference on real scenes). At each fragment $\mathbf{x}$ we trilinearly interpolate the features of every hash level, concatenate them, and decode with $f_\phi$. The grid has four levels with $4$-D features, over a base resolution that depends on the scene's extent (Appendix~\ref{sec:hyperparams}); $f_\phi$ has two hidden layers of width $16$.

\begin{figure}[h]
  \centering
  \includegraphics[width=0.78\linewidth]{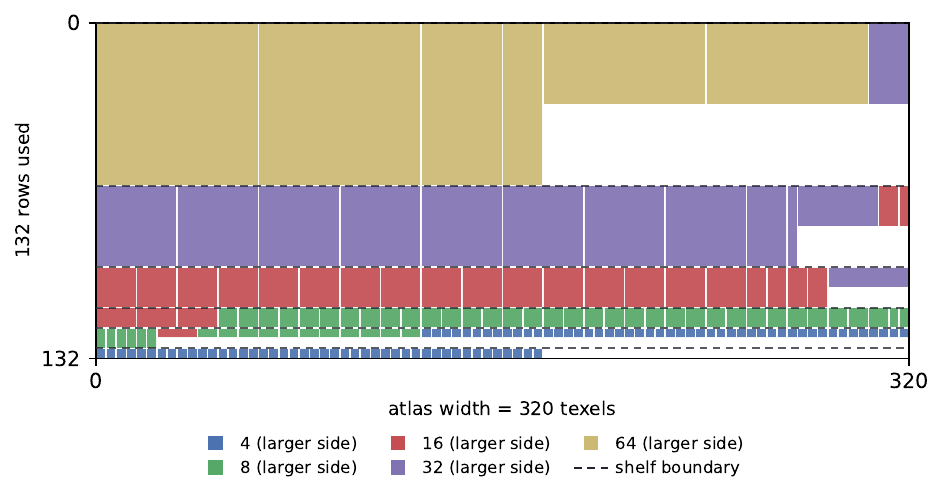}
  \caption{\textbf{Atlas packing.} Each surfel owns one rectangular tile with sides chosen by the rate--distortion probe ($4$ to $64$ texels); tiles are sorted by height and placed first-fit into full-width shelves, whose height is fixed by their first tile, hence the white gaps. The size mix is chosen for legibility; the full atlas is dominated by $64$-texel tiles and packs tighter.}
  \label{fig:atlas_packing}
\end{figure}

\textbf{Texture baking.} After training we evaluate the hash grid over every surfel and store the result in one RGB atlas, so inference needs no network. Each surfel $g$ receives an anisotropic resolution $\mathbf{r}_g = (r_x, r_y)$ matched to the Nyquist rate of the finest hash voxel $\delta$ along each tangent axis $a \in \{u, v\}$:
\begin{equation}
r_{g,a} = \mathrm{clamp}\left(2^{\lceil \log_2(4 e s_{g,a}/\delta) \rceil}, r_{\min}, r_{\max}\right),
\label{eq:nyquist}
\end{equation}
where $s_{g,a}$ is the surfel scale on axis $a$, $e$ is the UV half-extent in units of that scale, the factor $4$ combines the $2e$ span with Nyquist sampling, and $r_{\min}=4$, $r_{\max}=64$. Anisotropic rectangles avoid square padding on thin surfels. The rectangles are packed into a fixed-width atlas by Shelf-First-Fit-Decreasing (Fig.~\ref{fig:atlas_packing}), and each texel center is lifted through the surfel's tangent frame to the world position at which $\boldsymbol{\rho}$ is evaluated. Appendix~\ref{sec:mipmapping} discusses issues related to mipmapping the texture atlas.  

\textbf{Atlas layout and texel evaluation.}
\label{sec:bake_details}
Once the resolutions of Eq.~\ref{eq:nyquist} are determined, the per-surfel grids are stored in an atlas of dimensions $W \times H$ using a Shelf-First-Fit-Decreasing algorithm. Surfels are sorted descending by $r_y$, then $r_x$, and placed into the first available shelf. We fix $W=4096$ and grow $H$ greedily, rounding to a multiple of 64, and automatically increase $W$ if $H$ approaches the 65{,}536-row \texttt{cudaArray} limit.

At the center of each texel, the hash grid and MLP are evaluated. For a texel $(i,j)$ assigned to surfel $g$, local coordinates are computed using the texel-center convention:
\begin{equation}
u = \frac{i + 0.5}{r_x} 2e - e, \quad v = \frac{j + 0.5}{r_y} 2e - e.
\end{equation}
This mapping prevents boundary seams by aligning the baked evaluation with the hardware bilinear sampler. The local coordinates are lifted to world space via the surfel tangent frame $\mathbf{R}$:
\begin{equation}
\mathbf{x} = \mathbf{p}_g + u s_{g,u} \mathbf{R}_{:,0} + v s_{g,v} \mathbf{R}_{:,1}.
\end{equation}
The evaluated RGB vectors $\boldsymbol{\rho}(\mathbf{x})$ are written to the atlas in a batched pass that evaluates the hash grid and the fused MLP decoder over chunks of surfels, bounding peak memory during the bake.

\textbf{Support-aware extent.}
\label{sec:atlas_sizing}
The atlas covers a square of half-extent $e$ in surfel-scale units around each
surfel. At $e=4$ this patch reaches well beyond the kernel: the compact Beta
kernel is identically zero past $r=3\sigma$, so
on a representative indoor bake $50.5\%$ of all texels lie outside the support
even after a one-texel guard band for bilinear filtering, and are never
fetched. Baking over $e=3$ removes them. Because two thirds of the rectangles
are at the $r_{\max}$ cap, most of this saving is returned as a $33\%$ finer
texel pitch rather than as bytes.

\textbf{Rate--distortion allocation.}
Nyquist sizing follows the hash grid, not the contribution each surfel makes to
the rendered image. We instead measure per surfel what each smaller size of its
rectangle destroys, as the kernel- and opacity-weighted error of a
down- and up-sampled patch, weight it by the surfel's accumulated $\alpha T$
visibility across the training views, and let a single Lagrange multiplier fix
every rectangle's size; the multiplier is bisected to a distortion target. We
set that target to the error the BC7 codec already introduces, so the
allocation costs no more than the quantizer we use. The probe runs on the raw bake, which is band-limited by the hash grid, so halving it loses little; it keeps $45\%$ of the texels, and a single finetune of the smaller atlas recovers the quality. Probing the finetuned atlas instead keeps $60\%$, because the finetune adds detail beyond the grid's band and every fitted texel then looks necessary; that variant also costs a second finetune and over the 13 scenes gives $0.05$\,dB less PSNR and $0.002$ more LPIPS for $23\%$ more bytes. Table~\ref{tab:atlas_rd} compares the sized atlas with the full-extent shelf atlas under the same protocol: $0.44\times$ the bytes, higher PSNR and SSIM on every benchmark, lower LPIPS on two of three. The deepest cuts, \emph{playroom} at $-52\%$ bytes and \emph{room} at $-42\%$, are where LPIPS moves up by $0.002$--$0.010$, since the distortion target is MSE.

\begin{table}[h]
\centering
\small
\caption{\textbf{Atlas sizing.} Support-aware extent ($e=3$) with rate--distortion allocation probed on the raw bake, against the full-extent ($e=4$) shelf atlas. Both rows are BC7 bakes after one atlas finetune from the same teacher per scene, rendered by the inference renderer as in Table~\ref{tab:nvs_results}. Atlas sizes are decimal MB.}
\label{tab:atlas_rd}
\begin{tabular}{ll|ccc|r}
\toprule
\textbf{Benchmark} & \textbf{Atlas} & PSNR$\uparrow$ & SSIM$\uparrow$ & LPIPS$\downarrow$ & Atlas / scene \\
\midrule
\multirow{2}{*}{Mip-NeRF 360} & full ($e=4$) & 27.23 & 0.7892 & 0.2114 & 388 MB \\
 & sized & \textbf{27.31} & \textbf{0.7927} & \textbf{0.2079} & \textbf{173 MB} \\
\midrule
\multirow{2}{*}{Tanks \& Temples} & full ($e=4$) & 24.10 & 0.8507 & 0.1447 & 284 MB \\
 & sized & \textbf{24.11} & \textbf{0.8533} & \textbf{0.1406} & \textbf{132 MB} \\
\midrule
\multirow{2}{*}{Deep Blending} & full ($e=4$) & 29.87 & 0.8902 & \textbf{0.2176} & 228 MB \\
 & sized & \textbf{30.05} & \textbf{0.8975} & 0.2203 & \textbf{73 MB} \\
\bottomrule
\end{tabular}
\end{table}

\textbf{Quantization.} Instead of storing a dense FP16 RGB atlas, we apply Block Compression (BC7) to reduce the memory footprint. The residual range is estimated as the empirical $[\min, \max]$ with a $2\%$ margin over roughly $1.3$ million samples (64 random UV positions on each of 20{,}000 surfels), and the atlas is quantized uniformly with a single global $(\mathrm{offset}, \mathrm{scale})$ pair, written to the bake metadata and applied as one multiply-add per read by the rasterizer. Two details affect fidelity. First, range estimates of the form $\mu \pm k\sigma$ clip the heavy tails of the residual distribution. Clamping the bright peaks dims the render uniformly, so we use the full empirical range instead. Second, the scale is snapped so that a residual of zero coincides with an integer quantization level; otherwise the half-step rounding bias propagates through the outer $\mathrm{ReLU}$ of Eq.~\ref{eq:color} and systematically dims pixels near the activation threshold. The data is padded to RGBA to match the $4 \times 4$ block layout of BC7, which stores each block in 16 bytes, or 1 byte per texel. As Table~\ref{tab:quantization} shows, quantization introduces no perceivable quality loss at a large memory reduction.

\begin{table}[h]
\centering
\caption{Atlas storage formats. Bytes per texel and the resulting mean atlas size over the nine Mip-NeRF 360 scenes, in decimal MB.}
\label{tab:quantization}
\small
{%
\begin{tabular}{lccc}
\toprule
\textbf{Format} & \textbf{Bytes / texel} & \textbf{Mean atlas size} & \textbf{$\Delta$PSNR vs.\ neural} \\
\midrule
FP16 RGB (ref.) & 6 & $\sim$1039\,MB & -- \\
uint8 RGBA & 4 & $\sim$692\,MB & $\pm$0.01 dB \\
\textbf{BC7 RGBA (ours)} & \textbf{1} & \textbf{$\sim$173\,MB} & \textbf{$\pm$0.01 dB} \\
\bottomrule
\end{tabular}}
\end{table}

\textbf{Atlas finetuning.}
\label{sec:finetune}
Baking is a lossy projection of the neural residual: quantization, the Nyquist clamp on large surfels, and bake-time pruning all discard information, so a baked atlas can only lose quality relative to the hash grid that produced it. We recover this loss, and more, by treating the bake as an initialization. After baking, the atlas is dequantized into a single trainable FP16 texture and optimized for 3{,}000 iterations against the training images with the standard $\mathcal{L}_1$ + D-SSIM photometric loss. Geometry and view-dependent color stay frozen: each surfel's texels were evaluated at its bake-time world-space positions, so moving a surfel would invalidate the binding between its texels and the scene, and jointly training the view-dependent term is ill-posed because Eq.~\ref{eq:color} lets it cancel against the residual. A freeze ablation of our method (\emph{garden}, \emph{room}, \emph{bicycle}) confirms both effects: unfreezing geometry costs $0.18$\,dB and $0.009$ LPIPS even at a tenth of the stock learning rate, joint training at the full rate costs $2.7$\,dB, and unfreezing the view-dependent color alone is a tie ($0.01$\,dB).

The finetune renders through a differentiable variant of the inference rasterizer whose forward pass is the exact atlas fetch (rect transform, clamp, bilinear); the backward pass scatters the residual gradient into the four texels of each bilinear read, and the hash grid and MLP are never evaluated. Because only 5--7\% of texel rows receive gradient in any iteration, we apply a lazily allocated sparse Adam step over the touched rows with bf16 moments. This halves optimizer memory and, unlike a dense step, leaves untouched texels unchanged, avoiding the slow random walk that otherwise degrades the atlas once the coherent photometric signal is absorbed. The full stage costs a few minutes per scene, and primitive count, atlas bytes, and frame rate are unchanged. As Table~\ref{tab:finetune} shows, the finetuned atlas surpasses the neural model it was baked from, which we call the teacher, on PSNR and LPIPS on all three benchmarks. Two properties explain this. First, the atlas is free of hash collisions: the hash grid's fixed-size feature table forces distant scene locations to share table entries at its finest levels, so their gradients average into the same feature and high-frequency detail is lost, whereas every atlas texel is an independent parameter owned by a single surfel location. Second, the explicit texture holds $5$--$29\times$ ($13\times$ on average) the parameters of the network that produced it. Optimizing the texels directly therefore recovers detail the hash grid cannot represent.

\subsubsection{View-dependent Surfel Color}
Each surfel carries a soft spherical Voronoi model \citep{di2025spherical}: $K$ sites $s_k$ on the unit sphere with colors $c_k$ and temperatures $\tau_k$, blended as
\begin{equation}
f_{SV}(\mathbf{d}) = \sum_{k=1}^{K} w_k(\mathbf{d})\, c_k,
\quad w_k(\mathbf{d}) = \frac{\exp(-\tau_k \lVert s_k - \mathbf{d} \rVert)}{\sum_{k'=1}^{K}
\exp(-\tau_{k'} \lVert s_{k'} - \mathbf{d} \rVert)}.
\end{equation}
Smooth, learnable cell boundaries give a stable explicit parameterization, and $\tau_k$ sets the angular sharpness, so one surfel can hold both broad diffuse shifts and sharp highlights. The mixture is offset by $b_{SV}=0.5$ and gated by the inner $\mathrm{ReLU}$ of Eq.~\ref{eq:color}.

\subsection{Rendering}
\label{sec:hwraster}
For novel view synthesis, we replace the neural renderer with a 2D  rasterizer using hardware accelerated texture fetches. For each fragment in a surfel's footprint, the texture unit samples the BC7 block at the fragment's UV with bilinear filtering, and the dequantized residual $\hat{\boldsymbol{\rho}}$ enters the outer $\mathrm{ReLU}$ of Eq.~\ref{eq:color} before $\alpha$-compositing. When using hardware rasterization, our Vulkan port follows the vertex- and fragment-shader pipelines built for 3D Gaussians~\citep{niedermayr2024compressed,rotabulo2025hwraygs}: it precomputes per-surfel rendering parameters and generates depth keys in a compute pass to GPU sort visible surfels~\citep{niedermayr2024compressed} instead of per tile surfel sorting. Each surfel is then hardware-rasterized as a convex polygon circumscribing its exact screen support.

\textbf{Exact rational ray--surfel evaluation.}
\label{sec:conic}
A surfel is a flat rectangle in space. For each pixel, the renderer needs the point where the pixel's viewing ray hits that rectangle, in the surfel's own coordinates $(u,v)$. 2DGS solves this intersection from scratch at every pixel, with a cross product and two divisions. 3DGS works directly in image space and has no such step, so surfels pay a per-pixel cost that Gaussians do not. Most of that cost can leave the pixel loop.

 A flat surface seen in perspective maps to the image by a homography, so $(u,v)$ is a ratio of two linear functions of the pixel position, the same fact rasterizers exploit for perspective-correct texturing. Around the center of the surfel's screen-space bounding box,
\begin{equation}
u = u_0 + \frac{(\mathbf{J} \boldsymbol{\Delta})_u}{1 + \mathbf{d}^{\top} \boldsymbol{\Delta}}, \qquad
v = v_0 + \frac{(\mathbf{J} \boldsymbol{\Delta})_v}{1 + \mathbf{d}^{\top} \boldsymbol{\Delta}},
\label{eq:conic}
\end{equation}
where $\boldsymbol{\Delta}$ is the pixel offset from that center, $(u_0,v_0)$ the intersection there, $\mathbf{J}$ its rate of change per pixel, and $\mathbf{d}$ that of the perspective denominator. These eight numbers depend only on the surfel and the camera, so we compute them once per surfel and frame; per pixel, about 15 fused multiply--adds and two divisions become 10 and one, and a surfel then costs roughly what a Gaussian costs in 3DGS. The form is algebraically exact, and the renders match the 2DGS formulation to within $0.03$\,dB PSNR and $0.0002$ LPIPS on all 13 scenes (individual pixels differ where a floating-point difference near the alpha cutoff flips whether a fragment is drawn), at $32\%$ higher mean frame rate (Table~\ref{tab:ablation_kernel}). Pixels with a non-positive denominator look at the plane from behind the camera and are skipped. We derive this formulation below. Training keeps the 2DGS formulation, where the backward pass and hash-grid evaluation dominate.

\textbf{Derivation.}
\label{sec:FastGS_intersections}
A planar surfel seen through a pinhole camera projects to the screen by a homography: a $3\times3$ matrix $\mathbf{T}$ with rows $\mathbf{T}_u, \mathbf{T}_v, \mathbf{T}_w$ maps the surfel point $\mathbf{q} = (u, v, 1)$ to the screen point $(\mathbf{T}_u\!\cdot\mathbf{q},\, \mathbf{T}_v\!\cdot\mathbf{q})/(\mathbf{T}_w\!\cdot\mathbf{q})$. 2DGS inverts this per pixel: the pixel $(x,y)$ is hit by the surfel point that satisfies $\mathbf{T}_u\!\cdot\mathbf{q} = x\,\mathbf{T}_w\!\cdot\mathbf{q}$ and $\mathbf{T}_v\!\cdot\mathbf{q} = y\,\mathbf{T}_w\!\cdot\mathbf{q}$, two linear constraints $\mathbf{k}\cdot\mathbf{q} = 0$ and $\mathbf{l}\cdot\mathbf{q} = 0$ with $\mathbf{k} = x\,\mathbf{T}_w - \mathbf{T}_u$ and $\mathbf{l} = y\,\mathbf{T}_w - \mathbf{T}_v$. Their common solution is the cross product $\mathbf{p} = \mathbf{k}\times\mathbf{l}$, and dividing out the homogeneous scale gives $u = p_x/p_z$ and $v = p_y/p_z$. Per fragment this is two vector constructions, a cross product, and two divides: roughly 15 fused multiply--adds, two divides, and 56 bytes of shared-memory traffic.

The cross product does not have to be evaluated per fragment. Both $\mathbf{k}$ and $\mathbf{l}$ are affine in $(x,y)$, and their only bilinear term is $xy\,(\mathbf{T}_w\times\mathbf{T}_w)$, which vanishes because the cross product of a vector with itself is zero, so $\mathbf{p}$ is itself affine in the pixel. Each of $u$ and $v$ is therefore a ratio of two affine functions of the pixel, $u(x,y) = (a_0 + a_1 x + a_2 y)/(b_0 + b_1 x + b_2 y)$, the familiar form of perspective-correct interpolation. Writing that ratio around the center of the surfel's screen-space bounding box gives Eq.~\ref{eq:conic}: $(u_0,v_0)$ is the intersection at the center, $\mathbf{J}$ its Jacobian there, and $\mathbf{d}$ the gradient of the shared denominator, so the eight coefficients follow from one evaluation of $\mathbf{p}$ and its two partial derivatives per surfel. The form is the exact closed form, not a linearization. Where the denominator $1 + \mathbf{d}^{\top}\boldsymbol{\Delta}$ reaches zero the surfel plane is parallel to the ray, and beyond it the intersection lies behind the camera; the CUDA renderer discards those fragments per pixel, and the hardware port clips the polygon it draws by that line so they are never generated.

\textbf{Precision of the stored coefficients.} Precision determines whether the rational reformulation of Sec.~\ref{sec:conic} stays exact. The coefficients $(u_0, v_0)$, $\mathbf{J}$, opacity, and shape have small dynamic range and are stored as FP16, while $\mathbf{d}$ spans several orders of magnitude and stays FP32, since any error there scales the correction across the whole surfel. With this split, the closed-form intersection reproduces the reference formulation's quality to within $0.03$\,dB PSNR and $0.0002$ LPIPS on all 13 benchmark scenes (10 of 13 within $0.001$\,dB). The images are not byte-identical: rendering the same bake twice through one build is deterministic (PSNR between runs at the metric floor), while the two formulations differ by 59--66\,dB PSNR from each other, because floating-point reassociation near the alpha cutoff flips whether a fragment is drawn. The staging changes no arithmetic, only the shared-memory layout, and any rasterizer of this form could adopt it. It is also a different trade-off from packing the plane rows themselves at reduced precision, which routes FP16 inputs through the error-amplifying cross product and costs $0.05$--$0.7$\,dB; the closed-form intersection has no cross product, so the staging is safe here.

\subsection{Optimization}
Our hybrid representation already converges to a sparse set of primitives under standard density control. To further improve sparsity, we adopt the multi-view error criterion of FastGS \citep{ren2025fastgs}. We further propose a novel falloff regularizer on the Beta kernels, and we freeze the hash grid periodically to bound training time.

\textbf{Multi-view error criterion.}
\label{sec:FastGS_sparsity}
FastGS scores each primitive $i$ by the average number of high-error pixels in its footprint $\Omega_i$ over $K$ sampled training views,
\begin{equation}
s_d^i = \frac{1}{K} \sum_{j=1}^K \sum_{p \in \Omega_i} \mathbb{I}\big(\mathcal{M}^j_{\text{mask}}(p) = 1\big),
\end{equation}
where $\mathcal{M}^j_{\text{mask}}$ marks pixels with high photometric error in view $j$. The score gates densification: only surfels whose footprint covers high-error pixels are eligible, and surfels are pruned when their contribution to photometric accuracy is negligible. Which eligible surfel is cloned is then decided by the accumulated positional-gradient proxy $\partial\mathcal{L}/\partial\boldsymbol{\mu}_{2D}$. Because the hash grid is queried at the ray--surfel intersection, its backward pass feeds that proxy alongside the kernel-footprint term, and in a controlled test with a randomized grid and decoder the hash term accounts for $99.5\%$ of it. Flattening the kernel shrinks the footprint term further, so the signal that drives splitting is almost entirely the texture's.

\begin{figure}[t]
  \centering
  \includegraphics[width=\linewidth]{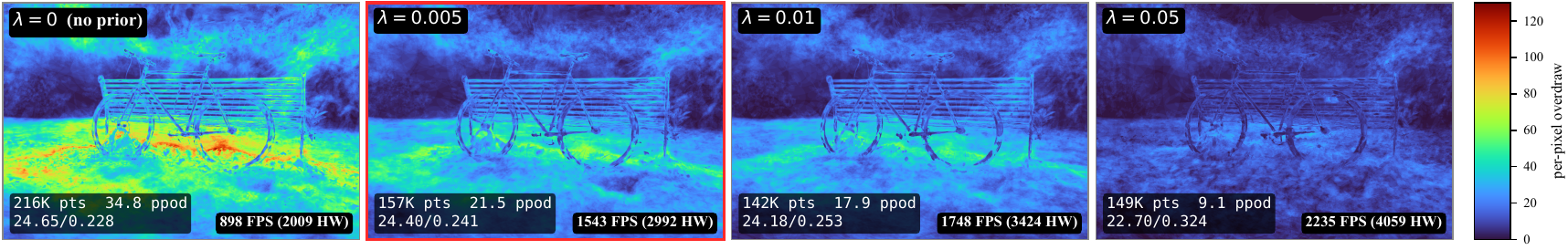}
  \vspace{-15pt}
  \caption{Falloff regularizer. For $\lambda = 0$, $0.005$, $0.01$, $0.05$ ($0.005$ is used in our experiments), per-pixel overdraw (ppod) is color coded from red (high) to blue (low). 
  At $0.05$, the count stops falling and quality collapses.}
  \label{fig:overdraw_lambda_gamma}
\end{figure}

\textbf{Falloff reduction.}
\label{sec:falloff}
A surviving Beta surfel can still spread influence beyond its geometric core through the soft tail of its kernel. We add a prior that pulls every shape parameter of Eq.~\ref{eq:beta-kernel} toward the flat end, $\mathcal{L}_{\beta} = \frac{1}{N}\sum_i \beta_i$, using the objective  $\mathcal{L} = \mathcal{L}_{\text{photo}} + \lambda\,\mathcal{L}_{\beta}$ with $\lambda = 0.005$. Balanced against the photometric term, the prior flattens those surfels whose soft tails the images do not require, from a Gaussian-like profile toward a hard-edged disk. This bounds the surfels' influence to their geometric extent and reduces pixel overdraw at inference (Fig.~\ref{fig:overdraw_lambda_gamma}). %
Appendix~\ref{sec:ablation_falloff_supp} varies $\lambda$ on all nine Mip-NeRF 360 scenes; $0.005$ is where the quality--speed trade-off bends.

\textbf{Training time.} In early training iterations, many fragments are rasterized and backpropagated through the hash grid and MLP. Thus, after 5{,}000 iterations we freeze both for nine of every ten iterations. The residual is close to converged by then, and freezing it gives the surfels a stable target to fit against at a fraction of the cost. Training averages 45 minutes per Mip-NeRF 360 scene, baking about 28 seconds at 12\,GB peak VRAM, and finetuning the atlas requires about four minutes.

\section{Experiments}
\label{sec:Results}
We evaluate on all scenes of Mip-NeRF 360~\citep{barron2022mip} and two scenes each from Tanks and Temples~\citep{knapitsch2017tanks} and Deep Blending~\citep{hedman2018deep}, against dense primitive-only methods (3DGS, 2DGS, Beta-Splatting), textured and hybrid methods (BBSplat, Nexels, Content-Aware Texturing, Textured Gaussians, NeST-Splatting), speed-optimized methods (Speedy-Splat, FastGS),
and primitive-budget variants of Beta-Splatting and Nexels. We report PSNR, SSIM, LPIPS, primitive count, storage, and frame rate. Every baseline is trained and rendered per scene with its authors' released code, configuration, and evaluation scripts; quality is recomputed 
by one shared metric script, which reproduces the authors' own numbers where reported. Only the frame rate is measured with our protocol: one harness calls each method's released render function on the held-out cameras, on the same idle NVIDIA RTX 5090 with \texttt{torch.cuda.Event} (50 warm-up, 400 timed frames, sort and preprocess included), with no method-specific optimization applied to any baseline. For timing the hardware rasterizer, 
we use Vulkan timestamp queries around the same three stages, with warm-up and timed frames recorded into a single submission. We report GPU time only, and measure inference on the quantized BC7 atlas rather than the unquantized texture optimized during atlas fine-tuning.
Quantization costs $0.001$--$0.002$ LPIPS and is included in every number we report. Speedy-Splat and FastGS
are timed with the screen-space bounding and tile-intersection optimizations in Appendix~\ref{sec:rastopt}, so the software rasterizer stack is matched where that comparison is made. Mip-NeRF 360 uses the standard downsampled images (outdoor at quarter, indoor at half resolution); the other two benchmarks use native resolution. ``Ours'' denotes the full pipeline: baked BC7 atlas, atlas finetune, and the closed-form intersection of Sec.~\ref{sec:conic}.

\begin{table}[t]
\centering
\caption{\textbf{Quality and performance statistics.} Rows are grouped into primitive-only, textured and hybrid, speed-optimized, and ours; (200K)/(40K)/(400K) are primitive-budget variants, and FastGS (Big) is that method's higher-quality configuration, scored with its own evaluation script, which agrees with the shared script to $0.03$\,dB on the base configuration. MB is each method's stored render payload in its own format and is not ranked. Compressed 3DGS is listed twice for the same reason we are: it has both a CUDA rasterizer and a hardware-rasterizing viewer (Sec.~\ref{sec:hw_baselines}). ``Ours'' uses the sized, baked, and finetuned atlas of Sec.~\ref{sec:atlas_sizing} (Table~\ref{tab:atlas_rd}); ``Ours (HQ)'' 
uses a Gaussian kernel with no falloff regularizer.
The last row reports statistics with the hardware rasterizer.}

\label{tab:nvs_results}
\resizebox{\textwidth}{!}{
\begin{tabular}{l|cccccc|cccccc|cccccc}
\toprule
& \multicolumn{6}{c|}{\textbf{Tanks and Temples}} & \multicolumn{6}{c|}{\textbf{Mip-NeRF 360}} & \multicolumn{6}{c}{\textbf{Deep Blending}} \\
\textbf{Method} & PSNR$\uparrow$ & SSIM$\uparrow$ & LPIPS$\downarrow$ & Points$\downarrow$ & MB & FPS$\uparrow$ & PSNR$\uparrow$ & SSIM$\uparrow$ & LPIPS$\downarrow$ & Points$\downarrow$ & MB & FPS$\uparrow$ & PSNR$\uparrow$ & SSIM$\uparrow$ & LPIPS$\downarrow$ & Points$\downarrow$ & MB & FPS$\uparrow$ \\
\midrule
3DGS                 & 23.74 & \third{0.857} & 0.169 & 1.57M & 372 & 259 & 27.54 & \second{0.819} & 0.215 & 2.73M & 646 & 211 & 29.71 & 0.911 & 0.238 & 2.47M & 584 & 216 \\
2DGS                 & 23.15 & 0.837 & 0.212 & 0.85M & 198 & 230 & 26.83 & 0.799 & 0.252 & 2.13M & 495 & 131 & 29.47 & 0.907 & 0.257 & 1.51M & 351 & 157 \\
Beta-Splatting (DBS) & \first{24.71} & \first{0.874} & \third{0.143} & 1.75M & 200 & 190 & \first{28.07} & \first{0.831} & \first{0.190} & 3.11M & 356 & 110 & 29.44 & 0.909 & 0.237 & 3.00M & 343 & 136 \\
DBS (200K)           & 23.38 & 0.835 & 0.214 & 0.20M & 23 & 524 & 26.66 & 0.773 & 0.294 & 0.21M & 23 & 469 & 29.64 & 0.902 & 0.276 & 0.20M & 23 & 520 \\
Compressed 3DGS      & 23.65 & 0.845 & 0.182 & 1.53M & 19 & 197 & 27.11 & 0.805 & 0.234 & 2.67M & 28 & 185 & 29.66 & 0.905 & 0.246 & 2.39M & 25 & 189 \\
\;\;+ HW rasterizer & \multicolumn{5}{c}{\emph{same model, up to 1.03\,dB lower}} & 1182 & \multicolumn{5}{c}{\emph{same model, up to 1.03\,dB lower}} & 984 & \multicolumn{5}{c}{\emph{same model, up to 1.03\,dB lower}} & 1172 \\
\midrule
BBSplat              & 23.70 & \third{0.857} & 0.150 & 0.30M & 226 & 80 & 26.80 & 0.787 & 0.236 & 0.24M & 176 & 48 & 29.34 & 0.905 & 0.257 & 0.16M & 111 & 49 \\
Nexels (40K)         & 22.25 & 0.794 & 0.211 & \first{0.04M} & 138 & 204 & 25.99 & 0.762 & 0.241 & \first{0.04M} & 138 & 119 & 29.34 & 0.898 & 0.231 & \first{0.04M} & 138 & 106 \\
Nexels (400K)        & 23.62 & 0.845 & 0.158 & 0.40M & 230 & 137 & 27.27 & 0.803 & \third{0.203} & 0.40M & 230 & 78 & \first{30.41} & \second{0.914} & \first{0.205} & 0.40M & 230 & 56 \\
Content-Aware Tex.   & 23.37 & 0.839 & 0.197 & 0.13M & 127 & 270 & 26.84 & 0.789 & 0.234 & 0.16M & 436 & 137 & 29.99 & \first{0.914} & 0.241 & 0.19M & 252 & 147 \\
Textured Gaussians   & 22.70 & 0.806 & 0.216 & \third{0.10M} & 3837 & 30 & 25.86 & 0.734 & 0.286 & \second{0.10M} & 3837 & 28 & 29.04 & 0.891 & 0.265 & \third{0.10M} & 3837 & 30 \\
NeST-Splatting       & 22.54 & 0.818 & 0.187 & 0.38M & 90 & 71 & 26.39 & 0.773 & 0.226 & 0.96M & 224 & 39 & 28.86 & 0.902 & 0.227 & 0.48M & 113 & 60 \\
\midrule
Speedy-Splat         & 23.44 & 0.825 & 0.240 & 0.18M & 43 & \third{1459} & 26.91 & 0.788 & 0.288 & 0.32M & 75 & \third{1259} & 29.62 & 0.907 & 0.268 & 0.25M & 59 & \third{1501} \\
FastGS               & 24.00 & 0.840 & 0.209 & 0.24M & 60 & 1243 & 27.44 & 0.797 & 0.259 & 0.39M & 98 & 1209 & 30.02 & 0.905 & 0.265 & 0.22M & 54 & 1451 \\
FastGS (Big)         & \third{24.36} & \third{0.857} & 0.176 & 0.54M & 135 & 959 & \third{27.80} & \second{0.819} & 0.216 & 1.16M & 288 & 890 & \third{30.31} & \third{0.912} & 0.244 & 0.65M & 161 & 1276 \\
\midrule
Ours (HQ)            & \second{24.51} & \second{0.861} & \first{0.135} & 0.15M & 347 & 1136 & \second{27.92} & 0.810 & \second{0.194} & 0.25M & 494 & 684 & \second{30.33} & 0.903 & \second{0.210} & 0.13M & 256 & 930 \\
Ours                 & 24.10 & 0.853 & \second{0.141} & \second{0.09M} & 171 & \second{2184} & 27.31 & 0.793 & 0.208 & \third{0.12M} & 228 & \second{1504} & 30.06 & 0.898 & \third{0.220} & \second{0.07M} & 104 & \second{1970} \\
\;\;+ HW rasterizer & \multicolumn{5}{c}{\emph{identical bake, quality at parity}} & \first{3016} & \multicolumn{5}{c}{\emph{identical bake, quality at parity}} & \first{2698} & \multicolumn{5}{c}{\emph{identical bake, quality at parity}} & \first{4413} \\
\bottomrule
\end{tabular}
}
\end{table}

\subsection{Novel View Synthesis}
Table~\ref{tab:nvs_results} reports the comparison. Our method is the fastest on all three benchmarks, at $7$--$9\times$ the frame rate of 3DGS and $8\times$ or more that of every textured-primitive baseline. Among the speed-optimized methods (Speedy-Splat, FastGS, and ours), ours has the best LPIPS on every one of the 13 scenes, by margins of up to $0.13$, with $2$--$4\times$ fewer primitives; against FastGS we are faster on every scene ($1.2$--$1.8\times$ on benchmark means) with PSNR near parity (mean $-0.13$\,dB on Mip-NeRF 360, $+0.10$\,dB on Tanks and Temples, $+0.04$\,dB on Deep Blending, within $0.7$\,dB on every scene). The remaining deficit lies in regions where sparsified Gaussians fit blurry low-frequency blobs that PSNR rewards and LPIPS penalizes (Fig.~\ref{fig:teaser} and Sec.~\ref{sec:qualitative_supp}). FastGS (Big) buys PSNR and SSIM with primitives: on Mip-NeRF 360 it is $0.49$\,dB above ours at $0.008$ worse LPIPS, with $9.5\times$ the primitives and $0.59\times$ the frame rate, and Ours (HQ) is above it in PSNR and LPIPS on all three benchmarks.

The quality-optimized Beta-Splatting holds the best PSNR and SSIM on Mip-NeRF 360 and Tanks and Temples and the best LPIPS on Mip-NeRF 360, at $19$--$26\times$ our primitive count and $11$--$14\times$ lower frame rates; on Tanks and Temples our LPIPS is $0.003$ below it, and on Deep Blending we surpass it on PSNR and LPIPS. Constrained to a comparable budget, untextured methods degrade sharply: Beta-Splatting at 200K primitives falls to $26.66$ PSNR and $0.294$ LPIPS on Mip-NeRF 360, against our $27.31$ and $0.208$ at $1.7\times$ fewer primitives and $3.2\times$ its frame rate. Appendix~\ref{sec:ablation_falloff_supp} (Table~\ref{tab:untextured_prior}) isolates the residual inside our own pipeline by switching only it off. Our stored payload, the BC7 atlas plus uncompressed geometry as every other row of the table is measured, averages 171, 228, and 104\,MB on Tanks and Temples, Mip-NeRF 360, and Deep Blending (per-scene sizes in Table~\ref{tab:ps_size}). Compressing the geometry for distribution reduces it to $13.7$\,MB per scene, which brings the Mip-NeRF 360 mean to $187$\,MB, and an optional block dictionary compresses the atlas a further $7.0\times$ for distribution (Appendix~\ref{sec:blockdict}).

\subsection{Qualitative Results}
\label{sec:qualitative_supp}
\begin{figure}[t]
  \centering
  \includegraphics[width=\linewidth]{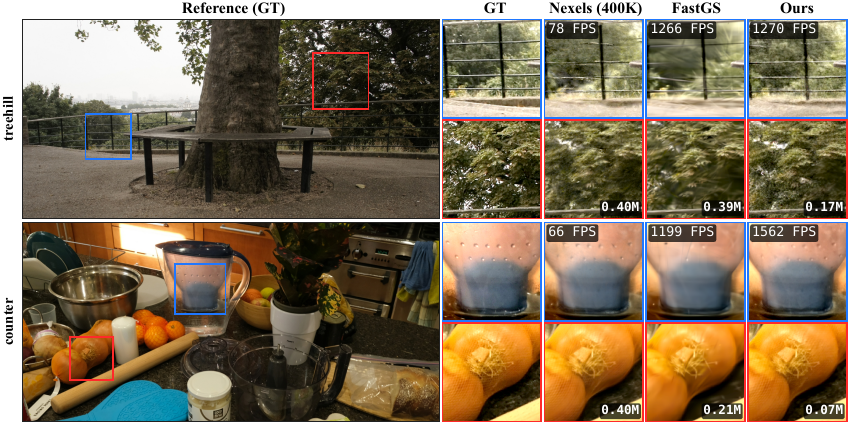}
  \caption{Qualitative results using software rasterization.}
  \label{fig:qualitative}
\end{figure}

\begin{figure}[tp]
  \centering
  \includegraphics[width=\linewidth]{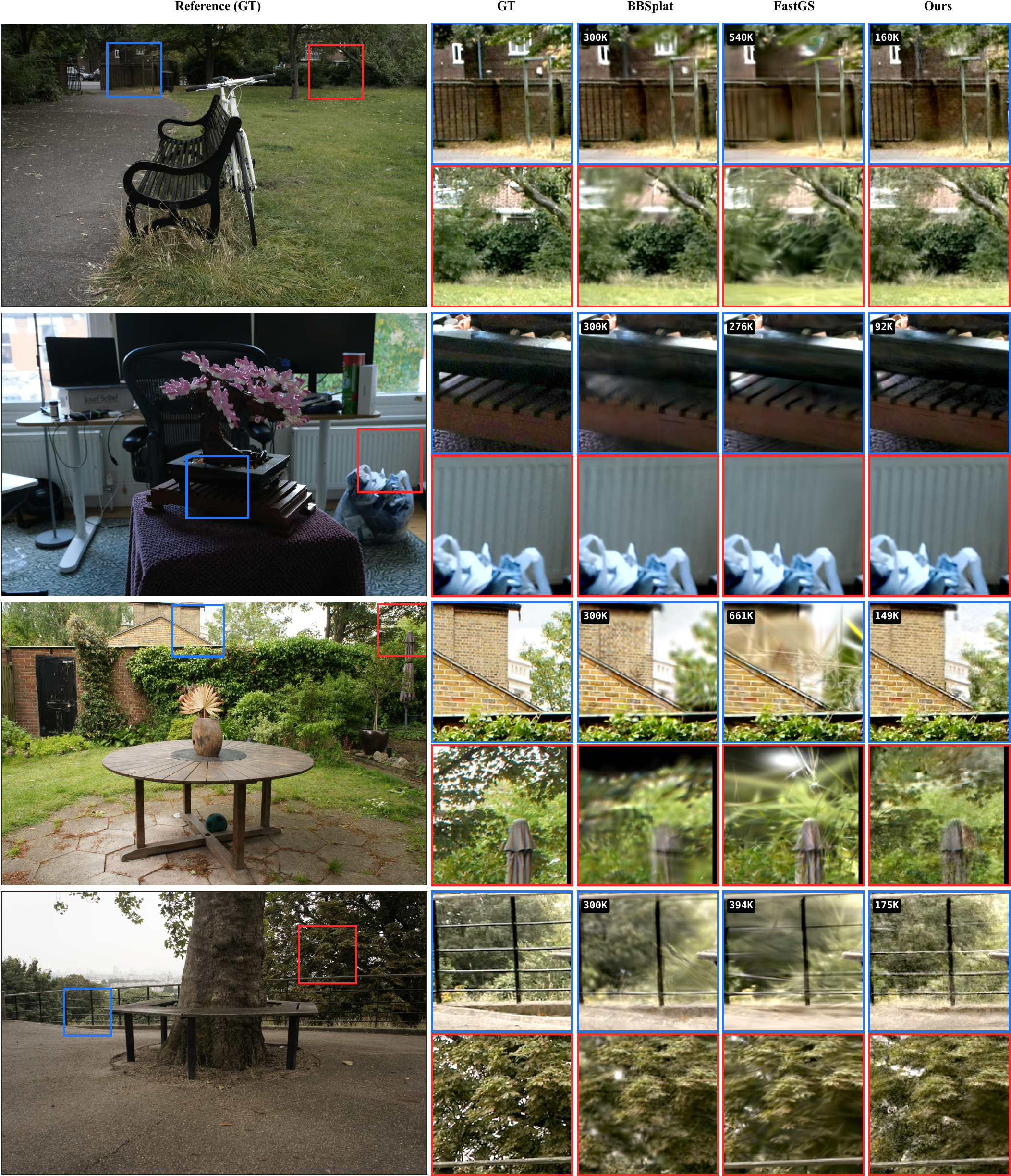}
  \caption{Qualitative comparisons of BBSplat, FastGS, and our method. Both BBSplat and FastGS struggle to represent background details that are captured well by our spatial hash-grid during training. Number of points, in thousands, for all methods.}
  \label{fig:qualitative_appendix}
\end{figure}

\begin{figure}[tp]
  \centering
  \includegraphics[width=\linewidth]{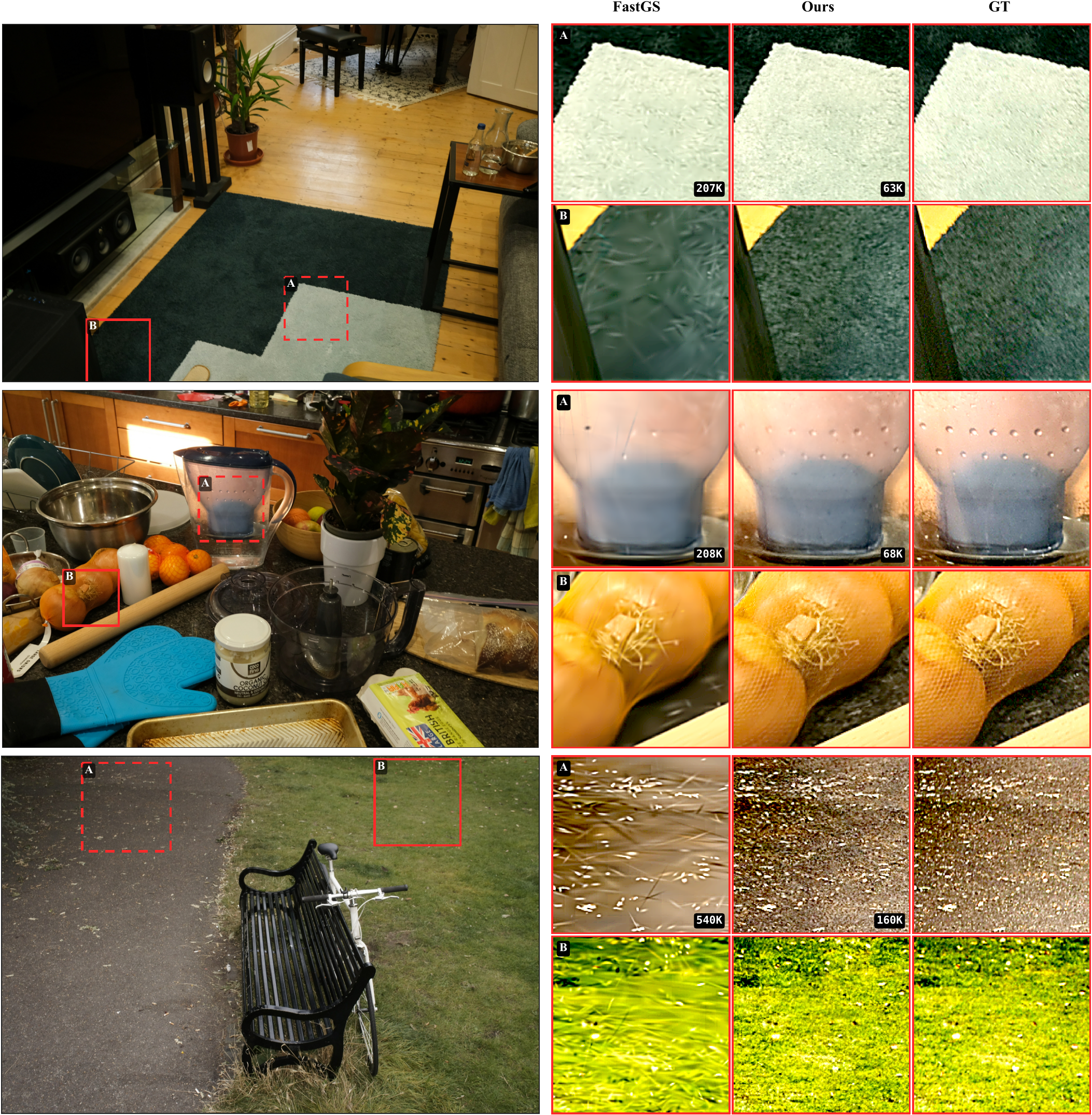}
  \caption{Scenarios where FastGS fails to represent foreground textures and falls back to blurry low-frequency Gaussians. Contrast-enhanced to highlight the textures. Number of points, in thousands, for FastGS and our method.}
  \label{fig:foreground-textures}
\end{figure}

Figure~\ref{fig:qualitative} compares renders of \emph{treehill} and \emph{counter} against Nexels (400K) and FastGS, and Figures~\ref{fig:qualitative_appendix} and~\ref{fig:foreground-textures} compare renders against BBSplat and FastGS on background and foreground detail.

\subsection{Ablations}
\label{sec:ablation_rast}
\textbf{Hardware rasterization.}
Quality-wise the GPU hardware rasterizer matches the CUDA software rasterizer on all 13 scenes, with at most 0.05\,dB and 0.0007 LPIPS deviation. The mean frame rates rise to 2698, 3016, and 4413 FPS on Mip-NeRF 360, Tanks \& Temples, and Deep Blending, respectively, $1.8\times$ over the CUDA implementation (Table~\ref{tab:nvs_results}). The same port gives 3DGS and FastGS $1.4\times$ and Speedy-Splat $1.7\times$ on average (Sec.~\ref{sec:hw_baselines}); on matched hardware rasterizers we remain $1.1$--$1.5\times$ faster than Speedy-Splat at lower LPIPS.
Because a hardware rasterizer never bins primitives into tiles and generates fragments only for pixels covered by the surfels, it needs significantly less computation than the software rasterizer
(Appendix~\ref{sec:rastopt}).
Specifically, 77\% of the fragments generated by the hardware rasterizer survive the support and opacity tests on average (53--86\% across the 13 scenes), against 17\% of the pixels when using software tile evaluation. The compact support of the Beta kernel makes the bound polygon tight: when using the Gaussian kernel (Ours HQ), nothing is culled by the support test and 29\% is culled by the opacity floor instead. Texturing costs 3--5\% of a frame versus 13--18\% in CUDA. Porting FastGS to the same hardware rasterizer, with its footprint crop enabled, gives it $1.41\times$ its CUDA frame rate on average over the 13 scenes ($0.96$--$2.10\times$), and $1.02\times$ without the crop; on \emph{garden} the port yields 1461 FPS for FastGS and 353 FPS for 3DGS, against 2842 FPS for ours. 
The performance gains are less for these methods because sorting consumes 44--47\% of their frame time against 25--39\% (mean 33\%) of ours.
As demonstrated in the comparison videos on the project page, sorting surfels globally instead of per tile does not introduce additional flickering in animations, even at low surfel counts. 

\label{sec:ablation_kernel}
\textbf{Kernel and falloff regularization.} Our regularizer acts on the Beta shape parameter and has no effect under a Gaussian kernel. In Table~\ref{tab:ablation_kernel} we compare
the unregularized Gaussian kernel, the Beta kernel without regularization, and the Beta kernel with regularization (ours). 
The Beta kernel leads to a $1.44$--$1.53\times$ increase in frame rate at a slight quality loss of $0.03$--$0.24$\,dB and $0.001$--$0.006$ LPIPS. 
The regularizer introduces an additional loss of $0.24$--$0.37$\,dB and $0.004$--$0.009$ LPIPS, at another $1.30$--$1.54\times$ increase in frame rate. %
On Mip-NeRF 360 the unregularized Beta kernel reaches 1752 FPS on the hardware rasterizer against 2698 for the regularized model, so the regularizer is what carries the throughput lead. 
For the kernel, the primitive count falls faster than the frame rate rises ($1.48\times$ versus $1.36\times$ on Mip-NeRF 360, software rasterizer), so its gain comes from fitting the scene with fewer primitives; for the regularizer, the frame rate rises faster than the count falls ($1.62\times$ versus $1.37\times$), so it also makes each surviving primitive cheaper, consistent with the overdraw reduction in Fig.~\ref{fig:overdraw_lambda_gamma}.

\begin{table}[t]
\centering
\caption{\textbf{Ablation on kernel regularization and ray-surfel intersection.} SW and HW, respectively, refer to the software and hardware rasterizer. All rows use the closed-form intersection of Sec.~\ref{sec:conic} except the one marked 2DGS intersection, which uses it on the same bake; that pair was timed in one session on the CUDA renderer, and the hardware rasterizer has no ray--surfel variant. Software frame rates of all rows were timed in one session on the bakes of Table~\ref{tab:nvs_results}.}
\label{tab:ablation_kernel}
\resizebox{\textwidth}{!}{
\begin{tabular}{l|ccccc|ccccc|ccccc}
\toprule
& \multicolumn{5}{c|}{\textbf{Tanks and Temples}} & \multicolumn{5}{c|}{\textbf{Mip-NeRF 360}} & \multicolumn{5}{c}{\textbf{Deep Blending}} \\
\textbf{Configuration} & PSNR$\uparrow$ & LPIPS$\downarrow$ & Points$\downarrow$ & \multicolumn{2}{c|}{FPS$\uparrow$} & PSNR$\uparrow$ & LPIPS$\downarrow$ & Points$\downarrow$ & \multicolumn{2}{c|}{FPS$\uparrow$} & PSNR$\uparrow$ & LPIPS$\downarrow$ & Points$\downarrow$ & \multicolumn{2}{c}{FPS$\uparrow$} \\
 & & & & SW & HW & & & & SW & HW & & & & SW & HW \\
\midrule
Gaussian kernel (regularizer inapplicable) & \first{24.51} & \first{0.135} & 0.15M & 1136 & \first{1509} & \first{27.92} & \first{0.194} & 0.25M & 684 & \first{1181} & \first{30.33} & \first{0.210} & 0.13M & 930 & \first{2053} \\
Beta kernel, no regularizer       & 24.37 & 0.137 & 0.11M & 1484 & 2314 & 27.68 & 0.200 & 0.17M & 929 & 1752 & 30.30 & 0.211 & 0.09M & 1194 & 2948 \\
\midrule
Beta kernel, with regularizer, 2DGS intersection & \multicolumn{3}{c}{\emph{identical bake}} & 1786 & -- & \multicolumn{3}{c}{\emph{identical bake}} & 1141 & -- & \multicolumn{3}{c}{\emph{identical bake}} & 1460 & -- \\
Beta kernel, with regularizer, closed-form intersection \emph{(ours)} & 24.10 & 0.141 & \first{0.09M} & \first{2184} & \first{3016} & 27.31 & 0.208 & \first{0.12M} & \first{1504} & \first{2698} & 30.06 & 0.220 & \first{0.07M} & \first{1970} & \first{4413} \\
\bottomrule
\end{tabular}}
\end{table}

\textbf{Atlas finetuning.} Table~\ref{tab:finetune} assesses the quality through the pipeline stages. Baking costs little PSNR but visible LPIPS relative to the neural teacher; finetuning recovers both and exceeds the teacher by $+0.24$ to $+0.47$\,dB PSNR and $0.025$--$0.026$ LPIPS on every benchmark, and SSIM on two of three.

\textbf{Closed-form intersection.} The last two rows of Table~\ref{tab:ablation_kernel} compare the closed-form intersection against the 2DGS formulation.
The saving is per fragment, so fragment-bound scenes gain most ($+47\%$ on \emph{stump} and \emph{flowers}) and scenes bound by sorting and binning gain least ($+20\%$ on \emph{train}).

\begin{table}[h]
\centering
\small
\caption{\textbf{Ablation on atlas finetuning.} Neural teacher, sized bake before finetuning, and finetuned atlas, all on the allocation of Table~\ref{tab:nvs_results}, so the last row equals its ``Ours'' row.
}
\label{tab:finetune}
\resizebox{\textwidth}{!}{
\begin{tabular}{l|ccc|ccc|ccc}
\toprule
& \multicolumn{3}{c|}{\textbf{Tanks and Temples}} & \multicolumn{3}{c|}{\textbf{Mip-NeRF 360}} & \multicolumn{3}{c}{\textbf{Deep Blending}} \\
\textbf{Stage} & PSNR$\uparrow$ & SSIM$\uparrow$ & LPIPS$\downarrow$ & PSNR$\uparrow$ & SSIM$\uparrow$ & LPIPS$\downarrow$ & PSNR$\uparrow$ & SSIM$\uparrow$ & LPIPS$\downarrow$ \\
\midrule
neural teacher & 23.82 & 0.8421 & 0.1662 & 26.84 & 0.7775 & 0.2328 & 29.82 & 0.8983 & 0.2457 \\
sized bake & 23.82 & 0.8400 & 0.1828 & 26.87 & 0.7771 & 0.2446 & 29.80 & \textbf{0.8990} & 0.2620 \\
baked + finetune & \textbf{24.10} & \textbf{0.8533} & \textbf{0.1407} & \textbf{27.31} & \textbf{0.7927} & \textbf{0.2079} & \textbf{30.06} & 0.8975 & \textbf{0.2203} \\
\bottomrule
\end{tabular}}
\end{table}

\subsection{Software versus Hardware Rasterization}
\label{sec:hw_baselines}
Sec.~\ref{sec:hwraster} reports our bakes on the fixed-function rasterizer. To separate what the hardware pipeline buys from what our primitives buy, we also run the baselines through a hardware rasterizer. FastGS and Speedy-Splat go through our Vulkan port. Compressed 3DGS~\citep{niedermayr2024compressed} has its own: its viewer sorts globally in a compute pass and draws each primitive as a quad through the graphics pipeline, as ours does. Its frames land $0.06$--$1.03$\,dB below its CUDA renders because it writes an 8-bit target, so Table~\ref{tab:nvs_results} takes its quality from the CUDA renders. Table~\ref{tab:hw_sw} sets each hardware rate against the software rate of Table~\ref{tab:nvs_results}. For 3DGS we port three scenes and compare against the bare CUDA rasterizer on the same bundle (Table~\ref{tab:hw_3dgs}).

\begin{table}[h]
\centering
\small
\caption{\textbf{Hardware port applied to 3DGS}, three scenes. Both columns are the same bundle through the same code path; every port is verified against the reference rasterizer to $\ge 54$\,dB on every view.}
\label{tab:hw_3dgs}
\begin{tabular}{l|rr|rr|c}
\toprule
scene & Gaussians & resolution & CUDA FPS & Vulkan FPS & speed-up \\
\midrule
garden & 4{,}144{,}122 & $1297\times840$ & 276 & 353 & $1.28\times$ \\
bonsai & 1{,}078{,}318 & $1559\times1039$ & 498 & 726 & $1.46\times$ \\
truck & 2{,}056{,}823 & $979\times546$ & 393 & 611 & $1.55\times$ \\
\midrule
\multicolumn{5}{l|}{mean} & $1.43\times$ \\
\bottomrule
\end{tabular}
\end{table}

\begin{table}[h]
\centering
\small
\caption{\textbf{Software versus hardware rasterization}, all 13 scenes. SW is the software rate of Table~\ref{tab:nvs_results}; HW is our Vulkan port for FastGS and Speedy-Splat, the authors' own hardware-rasterizing viewer for Compressed 3DGS, and our fixed-function path for our bakes. The rightmost column is the mean over the 13 scenes; for HW/SW it is the mean of the per-scene ratios.}
\label{tab:hw_sw}
\resizebox{\columnwidth}{!}{
\begin{tabular}{ll|ccccccccccccc|c}
\toprule
& & bicy. & flow. & gard. & stump & treeh. & room & count. & kitch. & bons. & train & truck & drj. & play. & mean \\
\midrule
\multirow{3}{*}{FastGS} & SW & 1212 & 1166 & 1146 & 1263 & 1266 & 1340 & 1199 & 1055 & 1229 & 1139 & 1348 & 1403 & 1498 & 1251 \\
 & HW & 1535 & 1427 & 1461 & 1626 & 1517 & 1834 & 1352 & 1011 & 1558 & 1865 & 2339 & 2713 & 3145 & 1799 \\
 & HW/SW & 1.27 & 1.22 & 1.27 & 1.29 & 1.20 & 1.37 & 1.13 & 0.96 & 1.27 & 1.64 & 1.74 & 1.93 & 2.10 & \textbf{1.41} \\
\midrule
\multirow{3}{*}{Speedy-Splat} & SW & 908 & 1196 & 1006 & 1062 & 1369 & 1479 & 1413 & 1414 & 1478 & 1508 & 1410 & 1390 & 1612 & 1327 \\
 & HW & 2124 & 1383 & 1701 & 1549 & 2195 & 2432 & 1957 & 1909 & 2286 & 3157 & 2527 & 2921 & 3071 & 2247 \\
 & HW/SW & 2.34 & 1.16 & 1.69 & 1.46 & 1.60 & 1.64 & 1.38 & 1.35 & 1.55 & 2.09 & 1.79 & 2.10 & 1.91 & \textbf{1.70} \\
\midrule
\multirow{3}{*}{Compressed 3DGS} & SW & 143 & 208 & 170 & 194 & 190 & 184 & 184 & 172 & 218 & 194 & 200 & 172 & 206 & 187 \\
 & HW & 797 & 1158 & 742 & 1049 & 1092 & 1068 & 956 & 792 & 1198 & 1211 & 1154 & 1035 & 1310 & 1043 \\
 & HW/SW & 5.57 & 5.57 & 4.36 & 5.41 & 5.75 & 5.80 & 5.20 & 4.60 & 5.50 & 6.24 & 5.77 & 6.02 & 6.36 & \textbf{5.55} \\
\midrule
\multirow{3}{*}{Ours} & SW & 1479 & 1263 & 1933 & 1419 & 1270 & 1966 & 1562 & 1354 & 1288 & 1996 & 2392 & 2109 & 1827 & 1681 \\
 & HW & 2988 & 2313 & 2842 & 3406 & 2607 & 3413 & 2294 & 1771 & 2650 & 2769 & 3262 & 4323 & 4503 & 3011 \\
 & HW/SW & 2.02 & 1.83 & 1.47 & 2.40 & 2.05 & 1.74 & 1.47 & 1.31 & 2.06 & 1.39 & 1.36 & 2.05 & 2.46 & \textbf{1.82} \\
\bottomrule
\end{tabular}}
\end{table}

A hardware rasterizer helps every method: $1.4\times$ on average for 3DGS and FastGS, $1.7\times$ for Speedy-Splat, and $1.82\times$ for our bakes. Compressed 3DGS gains $5.5\times$ because its CUDA path also re-decodes every primitive each frame. On matched hardware we are $1.7\times$ faster than FastGS, $1.3\times$ than Speedy-Splat, and $2.9\times$ than Compressed 3DGS on average over the 13 scenes.

\subsubsection{Temporal Stability of the Global Sort}
The hardware rasterizer sorts primitives once per frame, where the software renderer sorts per tile. A global sort can in principle pop where a per-tile sort does not. We render a 240-frame orbit through both and compare the second temporal difference (Table~\ref{tab:temporal}).

\begin{table}[h]
\centering
\small
\caption{\textbf{Popping under a 240-frame orbit}, mean absolute second temporal difference in 8-bit levels. The two renderers are indistinguishable.}
\label{tab:temporal}
\begin{tabular}{l|c|ccc}
\toprule
scene & SW vs HW & SW popping & HW popping & ratio \\
\midrule
bicycle & $57.32$\,dB & 35.328 & 35.315 & $1.000\times$ \\
garden & $56.92$\,dB & 36.579 & 36.569 & $1.000\times$ \\
\bottomrule
\end{tabular}
\end{table}

One limitation bounds what this measures. Adjacent orbit frames differ by 21 of 255 levels, so camera motion dominates the statistic and about half the pixels cross the popping threshold in both renderers. The table supports the comparative claim it was made for, not a claim about how much popping there is in absolute terms.

\FloatBarrier
\section{Conclusion, Limitations and Future Work}
We present a textured 2D surfel representation using a neural field that learns view-independent appearance during training and is finally baked into a compressed and finetuned texture atlas.
A closed-form ray--surfel intersection accelerates the software rasterizer by a third at unchanged quality, and hardware rasterization adds a further $1.4$--$2.2\times$. Our results demonstrate high speed and quality on common benchmarks, with the best perceptual quality among methods above 1000 FPS.

Compared to untextured methods our training is slower, due to hash-grid evaluations at high overdraw, and periodic freezing only partly hides that cost. To reduce the additional memory required by the texture atlas, a learned mapping from the trained hash grid to a fixed-size atlas is a natural next step. With the software rasterizer, sorting and binning bound additional fragment-level gains, and an overdraw loss driven by the per-pixel error is a natural extension of the falloff regularizer. The texture residual is view-independent, so view-dependent color is constant across a surfel, as it is for a Gaussian with spherical harmonics; finer view dependence falls to densification, and view-dependent texture residuals are a direction for future work. 

\subsection*{AI use statement}
We used generative AI assistants for language editing and polishing of the
manuscript text, for drafting figure- and table-generation scripts, and as a
coding assistant during the implementation of benchmarking and rendering
tooling. All AI-assisted text was reviewed and edited by the authors, and all
AI-assisted code was reviewed and tested by the authors. Research ideas, method
design, experiments, and analysis are the authors' own. We take responsibility
for the final content of this work.

\subsection*{Reproducibility statement}
All quantitative results use publicly available benchmarks (Mip-NeRF~360,
Tanks \& Temples, Deep Blending). The Experiments section specifies the full
measurement protocol: hardware, per-frame GPU timing methodology, warm-up and
frame counts, and the shared evaluation script applied to the renders of every
method. Appendix~\ref{sec:hyperparams} gives the full training configuration of
every stage, the learning rates, the densification thresholds, the bake and
allocation procedure, the finetuning settings, and the dataset splits and
resolutions; the appendix also reports per-scene results for the main
comparison and the ablations.
Interactive demos and videos are available on the project page, \url{https://nilkel.github.io/bitymi/}, where source code and pretrained checkpoints will be released.

\bibliographystyle{iclr2027_conference}
\bibliography{sample-base}
\newpage
\appendix
\section*{Appendix}
\section{Implementation Details and Hyperparameters}
\label{sec:hyperparams}
Our method is implemented on top of the NeST-Splatting codebase~\citep{zhang2025neural}, a 2D surfel rasterizer with a hash-grid and MLP appearance model. Settings not listed here are that codebase's defaults. One configuration is used for every scene; only the dataset entries of Table~\ref{tab:hp_data} differ. All timings are on one RTX 5090.

\subsection{Stage 1: neural model, 35k iterations from scratch}
\begin{table}[h]
\centering\small
\caption{Training configuration of the neural model that the atlas is baked from.}
\label{tab:hp_train}
{
\begin{tabular}{@{}p{0.22\linewidth}p{0.74\linewidth}@{}}
\toprule
\textbf{Component} & \textbf{Setting} \\
\midrule
primitive & 2D surfel, compact Beta kernel (Eq.~\ref{eq:beta-kernel}, support $k=3\sigma$), per-surfel shape; 2DGS screen-space low-pass filter \\
per-surfel color & spherical Voronoi basis, $K=7$ sites; base color $\mathrm{ReLU}(f_{SV}+0.5)$ \\
residual & multi-resolution hash grid, 4 levels $\times$ 4 features (the four finest of a six-level pyramid), $2^{21}$ entries per level, scene contraction to $[-5,5]$; fused MLP $16 \to 16 \to 16 \to 3$ with ReLU, queried at the ray--surfel intersection \\
fragment color & $\mathrm{ReLU}(\mathrm{ReLU}(f_{SV}+0.5) + \boldsymbol{\rho})$, Eq.~\ref{eq:color} \\
loss & $\mathcal{L}_1 + 0.2\,(1-\mathrm{SSIM})$ \\
falloff regularizer & $\lambda\,\mathcal{L}_\beta$ with $\lambda = 0.005$ (Sec.~\ref{sec:falloff}) \\
densification & multi-view criterion of Sec.~\ref{sec:FastGS_sparsity}, densify and prune every 500 iterations until 20k; gradient thresholds $1.5\mathrm{e}{-4}$ signed and $6\mathrm{e}{-4}$ absolute, importance threshold 30 views, loss threshold 0.1; late prune every 3k iterations from 20k to 30k, score threshold 0.9 \\
hash and MLP updates & every iteration until 5k, then every tenth iteration; frozen iterations reuse cached per-fragment residuals \\
learning rates & position $1.6\mathrm{e}{-4} \to 1.6\mathrm{e}{-6}$ exponential, opacity $0.05$, scale $5\mathrm{e}{-3}$, rotation $1\mathrm{e}{-3}$, shape $1\mathrm{e}{-3}$; SV sites $2\mathrm{e}{-3} \to 2\mathrm{e}{-4}$, SV colors $8\mathrm{e}{-4}$, SV temperature $6\mathrm{e}{-3}$ (sites and temperatures warm up over the first 1k iterations and are frozen for the last 3k); hash grid $2\mathrm{e}{-2}$, MLP $1\mathrm{e}{-3}$; Adam throughout \\
\bottomrule
\end{tabular}}
\end{table}

The two variants of Table~\ref{tab:ablation_kernel} differ from this in one setting each. \emph{Ours (HQ)} replaces the Beta kernel with a Gaussian, which leaves the falloff regularizer inapplicable because it acts on the Beta shape parameter. \emph{Ours (no reg.)} keeps the Beta kernel and sets $\lambda = 0$.

\subsection{Stage 2: bake and texel allocation}
No training happens in this stage.
\begin{enumerate}
\item \textbf{Bake.} Evaluate the trained hash grid and MLP on each surfel's tile over the kernel support, surfel UV in $[-3,3]^2$ at texel centres, up to $64\times64$ texels per surfel, in FP16.
\item \textbf{Rate--distortion probe.} For every surfel and every size on its halving ladder ($r_x \times r_y \to r_x/2 \times r_y/2 \to \dots$, each side clamped at $4$, then no texture), measure the importance-weighted distortion of the downsampled tile against the full-resolution bake.
\item \textbf{Allocation.} One global Lagrange multiplier, bisected so that the total importance-weighted distortion equals the BC7 floor; each surfel takes the size minimizing $D + \lambda\,\text{texels}$. Surfels assigned no texture render from their spherical Voronoi color alone.
\item \textbf{Re-bake} at the chosen sizes, shelf-packed into one atlas and BC7 compressed.
\end{enumerate}

\subsection{Stage 3: atlas finetune, 3k iterations}
\begin{table}[h]
\centering\small
\caption{Atlas finetuning configuration.}
\label{tab:hp_ft}
{
\begin{tabular}{@{}p{0.22\linewidth}p{0.74\linewidth}@{}}
\toprule
\textbf{Setting} & \textbf{Value} \\
\midrule
trained parameters & atlas texels only; geometry, view-dependent color and opacity frozen, hash grid and MLP discarded \\
loss & $\mathcal{L}_1 + 0.2\,(1-\mathrm{SSIM})$ against the training images \\
optimizer & Adam $(0.9, 0.999)$, learning rate $1\mathrm{e}{-3}$, lazy sparse updates on the texels touched by each view \\
renderer & differentiable form of the inference atlas fetch, bilinear, with the UV mapping used at inference \\
\bottomrule
\end{tabular}}
\end{table}

The finetuned atlas is re-quantized to BC7 for every number we report.

\subsection{Deployment format}
Web and mobile bundles store the atlas as a $4\times4$ block dictionary (K-means, $K=65{,}536$ codewords, each encoded once as BC7 or ASTC $4\times4$, with uint16 block indices). The viewer expands it into a regular hardware-compressed texture at load time, so rendering still performs one texture fetch per fragment. Surfel attributes use 8-bit min--max quantization, with positions in FP16.

\subsection{Datasets and evaluation}
\begin{table}[h]
\centering\small
\caption{Datasets, resolutions and the hash grid's base resolution.}
\label{tab:hp_data}
\resizebox{\textwidth}{!}{%
\begin{tabular}{llllr}
\toprule
\textbf{Dataset} & \textbf{Scenes} & \textbf{Split} & \textbf{Resolution} & \textbf{Hash base} \\
\midrule
Mip-NeRF 360, outdoor & bicycle, flowers, garden, stump, treehill & every 8th held out & \texttt{images\_4} & $2^{7}$ \\
Mip-NeRF 360, indoor & room, counter, kitchen, bonsai & every 8th held out & \texttt{images\_2} & $2^{8}$ \\
Tanks and Temples & train, truck & every 8th held out & native & $2^{8}$ \\
Deep Blending & drjohnson, playroom & every 8th held out & native & $2^{7}$ \\
\bottomrule
\end{tabular}}
\end{table}

PSNR, SSIM and LPIPS (VGG) are computed on the held-out views of the quantized BC7 model. Frame rates are measured on the held-out cameras at the evaluation resolution after warm-up, with GPU events around the full frame, so culling, sorting and rasterization are included.

\section{Rasterization Optimizations}
\label{sec:rastopt}
We incorporate the rasterization optimizations introduced by
Speedy-Splat~\citep{hanson2025speedy} and adopted in FastGS~\citep{ren2025fastgs}
to maximize rendering speed. We use SnugBox, a tight screen-space bounding box
derived directly from the conic equation, which provides numerical stability
under edge-on tilts. We also implement AccuTile, an analytic per-row ellipse
intersection during the duplicate-keys phase that emits keys only for tiles
the ellipse strictly overlaps.

\section{Atlas Compression and Mipmapping}
\subsection{Block-dictionary Compression for Transmission}
\label{sec:blockdict}
For distribution, the BC7 stream is optionally compressed with a block dictionary. The atlas is cut into $4\times4$ texel blocks, K-means clustering builds a codebook of $K=65{,}536$ block centroids, each centroid is encoded as one 16-byte BC7 block, and every atlas block stores a 16-bit codebook index. The codebook occupies 1\,MB and the indices 2 bytes per block, a $7.0\times$ reduction of the atlas payload (173\,MB to 24.7\,MB on average for Mip-NeRF 360), and the complete web bundle, dictionary atlas with 8-bit surfels (FP16 positions) and cameras, is 38.4\,MB. At load time, the indices are gathered back into a dense BC7 stream, so the renderer is unchanged and keeps a single hardware bilinear fetch per fragment. The dictionary therefore reduces download size, while GPU memory and frame time are those of the raw BC7 atlas.

\subsection{Mipmapping the Atlas}
\label{sec:mipmapping}
A distant surfel covers few pixels but keeps its full tile, so one pixel can span several texels and the single bilinear fetch undersamples the atlas. Our viewer can load a mip chain for this case: each level is a box filter of level~0, computed in the quantized domain and compressed like level~0, and four extra levels add $33\%$ to the atlas bytes. The level is chosen once per surfel from the texel footprint of a one-pixel step, which the closed-form intersection of Sec.~\ref{sec:conic} already provides, so rendering keeps one bilinear fetch per fragment. Because the finetune optimizes texels as they are read at level~0, filtering to the full footprint costs $0.32$\,dB and $0.045$ LPIPS on \emph{garden}; choosing one level sharper leaves quality at parity ($27.20$ / $0.8417$ / $0.1270$ against $27.06$ / $0.8405$ / $0.1247$ without mips) and the frame rate unchanged on the RTX~5090, where texture reads are a small share of the frame. Mipmapping is therefore off by default, and every number in this paper is rendered at level~0.

\section{Additional Ablations}
These studies support design choices that are not among the main claims: the sign of
the texture residual, the strength of the falloff prior, and the per-primitive
view-dependent color representation.

Two controls are absent by construction. A primitive-matched comparison between kernels is not available in this regime: the multi-view criterion of Sec.~\ref{sec:FastGS_sparsity} sets the primitive count from the error maps and exposes no budget, so forcing the Gaussian-kernel model down to the count of the regularized one means changing the densification thresholds, which changes where the scene is densified and therefore more than the kernel. We report each configuration at the count it converges to, next to that count. Training explicit per-surfel textures from the start, without the hash grid, is not a usable baseline either: textures would have to be allocated for every surfel alive during densification, before pruning has removed the redundant ones, and their texels, gradients and optimizer state would grow with every clone at the overdraw of the unpruned model. The hash grid caps that cost during training, and the atlas is allocated once, for the surfels that survive.

\subsection{Ablation: Signed versus Positive-only Texture}
\label{sec:ablation_signed}
\begin{table}[h]
\centering
\small
\caption{\textbf{Signed versus positive-only residual}, full pipeline (35k training, BC7 bake, 3k atlas finetune) on a fixed allocation, so the two activations are matched; quality from the CUDA renderer, frame rates from the CUDA (SW) and hardware (HW) rasterizers. Ablation arms use the previous shelf allocation, so absolute values differ from Table~\ref{tab:nvs_results}; the comparison is within each pair. $\Delta$ is positive-only minus signed.}
\label{tab:ablation_signed}
\resizebox{\columnwidth}{!}{
\begin{tabular}{l|ccc|ccc|r|rr|rr}
\toprule
& \multicolumn{3}{c|}{\textbf{Signed (ours)}} & \multicolumn{3}{c|}{\textbf{Positive-only}} & $\Delta$ LPIPS & \multicolumn{2}{c|}{SW FPS} & \multicolumn{2}{c}{HW FPS} \\
Scene & PSNR$\uparrow$ & SSIM$\uparrow$ & LPIPS$\downarrow$ & PSNR$\uparrow$ & SSIM$\uparrow$ & LPIPS$\downarrow$ & & signed & positive & signed & positive \\
\midrule
bicycle & 24.44 & 0.7137 & 0.2383 & 24.45 & 0.7134 & 0.2391 & $+0.0008$ & 1458 & 1444 & 2797 & 2894 \\
bonsai & 32.62 & 0.9380 & 0.1704 & 31.99 & 0.9325 & 0.1800 & $+0.0096$ & 1278 & 1358 & 2608 & 2774 \\
counter & 29.32 & 0.9010 & 0.1833 & 29.07 & 0.8977 & 0.1922 & $+0.0089$ & 1536 & 1661 & 2219 & 2437 \\
flowers & 20.78 & 0.5578 & 0.3200 & 20.69 & 0.5544 & 0.3224 & $+0.0024$ & 1258 & 1349 & 2289 & 2430 \\
garden & 27.03 & 0.8376 & 0.1288 & 26.96 & 0.8389 & 0.1286 & $-0.0002$ & 1904 & 2004 & 2785 & 2865 \\
kitchen & 31.47 & 0.9190 & 0.1233 & 30.83 & 0.9162 & 0.1299 & $+0.0066$ & 1337 & 1451 & 1732 & 2023 \\
room & 31.56 & 0.9193 & 0.1891 & 31.07 & 0.9138 & 0.2019 & $+0.0128$ & 1908 & 1951 & 3365 & 3501 \\
stump & 25.76 & 0.7228 & 0.2632 & 25.13 & 0.6585 & 0.3281 & $+0.0649$ & 1402 & 1468 & 3261 & 3515 \\
treehill & 22.56 & 0.6015 & 0.2937 & 22.52 & 0.5751 & 0.3596 & $+0.0659$ & 1263 & 1302 & 2601 & 2844 \\
\textbf{Mip-NeRF 360 mean} & 27.28 & 0.7901 & 0.2122 & 26.97 & 0.7778 & 0.2313 & $+0.0191$ & 1483 & 1554 & 2629 & 2809 \\
\midrule
drjohnson & 29.57 & 0.8910 & 0.2317 & 29.19 & 0.8931 & 0.2432 & $+0.0115$ & 2184 & 2299 & -- & -- \\
playroom & 30.39 & 0.8928 & 0.2082 & 30.38 & 0.8933 & 0.2177 & $+0.0095$ & 1878 & 1924 & -- & -- \\
\textbf{Deep Blending mean} & 29.98 & 0.8919 & 0.2199 & 29.79 & 0.8932 & 0.2304 & $+0.0105$ & 2031 & 2112 & -- & -- \\
\bottomrule
\end{tabular}}
\end{table}
The residual enters inside the outer $\mathrm{ReLU}$ of Eq.~\ref{eq:color}, so it may subtract from the per-primitive base color as well as add to it. Restricting it to add only, $\mathrm{ReLU}(\text{base}+b_{SV}) + \mathrm{ReLU}(\boldsymbol{\rho})$, gives a strictly smaller function class: any non-negative correction is reproduced by the signed model, while the converse fails wherever the texture must darken the base. A non-negative texture would be simpler to store and to read in a viewer, so Table~\ref{tab:ablation_signed} measures the cost of the restriction through the full pipeline, training, bake, and atlas finetune, on the nine Mip-NeRF 360 scenes and the two Deep Blending scenes, with identical commands apart from the activation.

The signed residual is better in LPIPS on 10 of the 11 scenes, the exception being \emph{garden}, where positive-only is lower by $0.0002$ and in PSNR on 10 of 11, by $0.31$\,dB and $0.019$ LPIPS on Mip-NeRF 360 and $0.19$\,dB and $0.011$ LPIPS on Deep Blending. Indoor scenes lose about $0.5$\,dB and $0.01$ LPIPS consistently; on \emph{stump} and \emph{treehill} the positive-only model loses $0.065$ LPIPS, and the loss is already present in the neural model before baking, so it comes from training rather than from the atlas. The activation adds no evaluation cost: the positive-only model renders $5$--$7\%$ faster only because it converges to about $2.5\%$ fewer primitives, since with a non-negative residual about half its channels are gated closed and the hash gradient that drives densification is correspondingly weaker.

\subsection{Ablation: Falloff Penalty Strength}
\label{sec:ablation_falloff_supp}

\textbf{The prior without the texture.} The prior is cheap only because the texture carries the colour variation inside a primitive. To measure this we train the two Table~\ref{tab:nvs_results} teacher recipes, prior off and prior at $\lambda = 0.005$, with and without the hash residual on \emph{bicycle}, \emph{garden}, and \emph{room}; the untextured arms disable the hash grid for the whole run, so the residual is exactly zero and contributes nothing to densification, and everything else (surfels, Beta kernel, spherical Voronoi colour, densification, low-pass filter, and the prior) is identical. Table~\ref{tab:untextured_prior} reports the neural test render at 35k iterations and the mean number of blended fragments per pixel. Switching the prior on costs the textured model $0.33$\,dB and $0.014$ LPIPS for $0.59\times$ the overdraw; the untextured model pays $0.53$\,dB and $0.026$ LPIPS, about twice as much on every one of the three scenes, for a smaller overdraw reduction ($0.65\times$). At matched overdraw the gap is larger still: on \emph{bicycle} the textured model with the prior (20.2 blends per pixel) is $0.125$ LPIPS better than the untextured model without any prior (19.1). Without a texture the kernel falloff is still interpolating colour between neighbouring primitives, and flattening it removes that; with a texture it only sets coverage. BBSplat and Textured Gaussians learn per-primitive alpha textures, which are arbitrary learned falloffs, the opposite design, and one that raises overdraw rather than lowering it.

\begin{table}[h]
\centering
\small
\caption{\textbf{The falloff prior with and without the texture residual.} Neural model at 35k iterations, identical recipe apart from the hash residual and $\lambda$; blends per pixel is the mean number of ray--surfel intersections per pixel over the test views.}
\label{tab:untextured_prior}
\resizebox{\columnwidth}{!}{
\begin{tabular}{ll|ccc|rr|ccc|rr}
\toprule
& & \multicolumn{5}{c|}{\textbf{textured}} & \multicolumn{5}{c}{\textbf{untextured}} \\
Scene & $\lambda$ & PSNR$\uparrow$ & SSIM$\uparrow$ & LPIPS$\downarrow$ & $N$ & blends/px & PSNR$\uparrow$ & SSIM$\uparrow$ & LPIPS$\downarrow$ & $N$ & blends/px \\
\midrule
bicycle & 0 & 24.30 & 0.7085 & 0.2534 & 214{,}895 & 33.9 & 23.58 & 0.6202 & 0.3964 & 178{,}399 & 19.1 \\
bicycle & 0.005 & 24.03 & 0.6900 & 0.2712 & 160{,}835 & 20.2 & 23.20 & 0.5844 & 0.4315 & 131{,}275 & 11.3 \\
garden & 0 & 27.08 & 0.8413 & 0.1375 & 207{,}616 & 34.3 & 25.59 & 0.7553 & 0.2961 & 176{,}094 & 17.1 \\
garden & 0.005 & 26.72 & 0.8290 & 0.1522 & 148{,}803 & 20.1 & 25.07 & 0.7225 & 0.3213 & 145{,}622 & 12.2 \\
room & 0 & 30.97 & 0.9142 & 0.2056 & 90{,}240 & 25.8 & 30.78 & 0.8965 & 0.2766 & 88{,}301 & 19.0 \\
room & 0.005 & 30.62 & 0.9073 & 0.2148 & 63{,}164 & 15.4 & 30.10 & 0.8817 & 0.2949 & 65{,}394 & 11.9 \\
\midrule
cost of the prior & & $-0.33$\,dB & & $+0.0139$ & $\times 0.72$ & $\times 0.59$ & $-0.53$\,dB & & $+0.0262$ & $\times 0.77$ & $\times 0.65$ \\
\bottomrule
\end{tabular}}
\end{table}

Table~\ref{tab:ablation_kernel} switches the prior off entirely; here we vary its weight $\lambda$ on all nine Mip-NeRF 360 scenes with the full pipeline at each value (Table~\ref{tab:ablation_falloff}; Figure~\ref{fig:overdraw_lambda_gamma} shows \emph{bicycle} at four values, and Table~\ref{tab:falloff_bake_all_scenes} lists every scene). Flatter kernels bound each primitive's blending footprint, so overdraw and primitive count fall together. $\lambda = 0.005$ is where the curve bends: the first step gives $1.53\times$ the hardware frame rate for $0.008$ LPIPS, the next $1.18\times$ for $0.011$, and the last $1.20\times$ for $0.068$, so each further step costs about four and then twenty times as much quality per unit of speed. Every scene follows the pattern: at $0.005$ the hardware frame rate rises by $1.46$--$1.69\times$ for $0.001$--$0.021$ LPIPS. At $0.05$ the discs are fully hard-edged and the prior works against itself, since densification has to add surfels to cover holes: 8 of nine scenes keep more primitives than at $0.01$, and PSNR drops by $2.1$\,dB.

\begin{table}[h]
\centering
\small
\caption{\textbf{Falloff-regularization ablation}, mean over the nine Mip-NeRF 360 scenes, full pipeline (trained, baked, texture-finetuned) on a fixed allocation, so the values of $\lambda$ are matched; hardware rasterizer on an RTX 5090. Ablation arms use the previous shelf allocation, so absolute values differ from Table~\ref{tab:nvs_results}; the comparison is within each pair. The prior $\lambda\,\mathcal{L}_\beta$ at each $\lambda$.}
\label{tab:ablation_falloff}
\resizebox{\columnwidth}{!}{
\begin{tabular}{l|rccc|rrr}
\toprule
$\lambda$ & $N$ & PSNR$\uparrow$ & SSIM$\uparrow$ & LPIPS$\downarrow$ & HW FPS$\uparrow$ & FPS $\times$ & $\Delta$ LPIPS \\
\midrule
0 (off) & 168{,}155 & 27.67 & 0.8013 & 0.2050 & 1732 & 1.00 & -- \\
\textbf{0.005} \emph{(ours)} & 121{,}445 & 27.26 & 0.7888 & 0.2131 & 2642 & 1.53 & +0.0081 \\
0.01 & 111{,}712 & 26.90 & 0.7778 & 0.2238 & 3123 & 1.80 & +0.0188 \\
0.05 & 124{,}079 & 24.82 & 0.7082 & 0.2921 & 3744 & 2.16 & +0.0870 \\
\bottomrule
\end{tabular}}
\end{table}

\subsection{Ablation: Per-primitive Features}
We compare the per-primitive view-dependent color representations on \emph{bonsai}: spherical harmonics (SH, degree 3), spherical Betas (SB, two lobes), and the spherical Voronoi model we use (SV, seven sites), with the same training recipe and both atlas allocations (Table~\ref{tab:ablation_spherical}). SV is $1.06$\,dB above SH and $0.64$\,dB above SB at the same primitive count, both read from the sized atlas, and has the best LPIPS by $0.003$. SB stores $15$ parameters per surfel against $49$ and costs $0.64$\,dB; at these sizes the atlas is an order of magnitude larger than the per-surfel parameters, so the saving rarely matters. Frame rate barely depends on the feature, within $4\%$ on the hardware rasterizer, since the color evaluation is a small share of the frame; SB is about $5\%$ slower in software because of its per-fragment lobe evaluation. The sized atlas cuts the bytes to about $45\%$ and improves LPIPS by $0.004$ for every feature, so the allocation does not depend on the color representation.

\begin{table}[h]
\centering
\small
\caption{\textbf{Per-primitive view-dependent color ablation} (\emph{bonsai}, full pipeline: trained, baked, texture-finetuned). Only the feature type differs; each is reported with the full-extent shelf atlas and with the sized atlas (here the earlier two-finetune allocation). \emph{Params} counts render-time per-surfel fields; atlas sizes are decimal MB.}
\label{tab:ablation_spherical}
\resizebox{\columnwidth}{!}{
\begin{tabular}{l|rr|lr|ccc|rr}
\toprule
\textbf{Feature} & $N$ & Params & Atlas & MB & PSNR$\uparrow$ & SSIM$\uparrow$ & LPIPS$\downarrow$ & SW FPS$\uparrow$ & HW FPS$\uparrow$ \\
\midrule
SH (deg 3) & 94{,}161 & 48 & full ($e=4$) & 254 & 31.57 & 0.9349 & 0.1724 & 1263 & 2551 \\
 & & & sized & 115 & 31.57 & 0.9353 & 0.1687 & 1268 & 2612 \\
SB ($K{=}2$ lobes) & 92{,}965 & \textbf{15} & full ($e=4$) & 254 & 31.97 & 0.9353 & 0.1732 & 1201 & 2511 \\
 & & & sized & 113 & 31.99 & 0.9361 & 0.1694 & 1213 & 2499 \\
SV ($K{=}7$ sites) \emph{(ours)} & \textbf{92{,}281} & 49 & full ($e=4$) & 243 & \textbf{32.62} & \textbf{0.9380} & \textbf{0.1704} & 1278 & 2608 \\
 & & & sized & 110 & \textbf{32.63} & \textbf{0.9385} & \textbf{0.1662} & 1291 & 2606 \\
\bottomrule
\end{tabular}}
\end{table}

\section{Additional Measurements}
\label{sec:extra_measurements}
This section collects controls and baselines measured after the main results were fixed. All frame rates are on the same RTX 5090 with an otherwise idle GPU, under the protocol of Sec.~\ref{sec:Results}.

\subsection{Is the Finetune Just Longer Training?}
The atlas finetune runs 3{,}000 iterations after the teacher converges, so the gain could be training length rather than the texels. We continue each teacher for the same 3{,}000 iterations with the hash grid unfrozen, densification off, and the learning-rate schedule continued rather than restarted (Table~\ref{tab:teacher_control}).

\begin{table}[h]
\centering
\small
\caption{\textbf{Teacher control, 13 scenes.} Neural test render at the end of each run; the last row is the ``Ours'' row of Table~\ref{tab:nvs_results}. Continuing the teacher explains none of the PSNR gain and about a third of the LPIPS gain.}
\label{tab:teacher_control}
\begin{tabular}{l|ccc}
\toprule
& PSNR$\uparrow$ & SSIM$\uparrow$ & LPIPS$\downarrow$ \\
\midrule
teacher at 35k & 26.84 & 0.8060 & 0.2245 \\
teacher $+$ 3k more, schedule continued & 26.78 & 0.8089 & 0.2165 \\
atlas finetune, same 3k on the texels & \textbf{27.24} & \textbf{0.8181} & \textbf{0.1995} \\
\bottomrule
\end{tabular}
\end{table}

\subsection{Seed Variance of the Signed Residual}
Table~\ref{tab:ablation_signed} reports one seed per scene, and its mean gap is carried by \emph{stump} and \emph{treehill}. Table~\ref{tab:seed_signed} repeats both arms on those two scenes with three seeds each.

\begin{table}[h]
\centering
\small
\caption{\textbf{Three seeds of the signed and positive-only arms}, baked LPIPS on the two scenes where the gap is largest. The direction holds on every seed, but the magnitude does not: the positive-only arm's spread across seeds is an order of magnitude larger.}
\label{tab:seed_signed}
\begin{tabular}{l|ccc|ccc}
\toprule
& \multicolumn{3}{c|}{\emph{stump}} & \multicolumn{3}{c}{\emph{treehill}} \\
seed & signed & positive & $\Delta$ & signed & positive & $\Delta$ \\
\midrule
0 (reported) & 0.2493 & 0.3280 & $+0.0787$ & 0.2874 & 0.3564 & $+0.0690$ \\
1 & 0.2451 & 0.2507 & $+0.0056$ & 0.2813 & 0.2884 & $+0.0071$ \\
2 & 0.2474 & 0.3141 & $+0.0668$ & 0.2856 & 0.3662 & $+0.0806$ \\
\midrule
mean & 0.2473 & 0.2976 & $+0.0504$ & 0.2848 & 0.3370 & $+0.0522$ \\
spread & 0.0042 & 0.0773 & & 0.0061 & 0.0778 & \\
\bottomrule
\end{tabular}
\end{table}

The signed arm wins all six seed pairs, so the ordering is not seed-dependent, but the reported gaps are the worst and middle of three. The result to read from this table is the spread: the positive-only constraint does not cost a fixed amount of quality, it makes the optimization unreliable.

\subsection{What May Train During the Finetune}
The atlas finetune trains texels only, with geometry and view-dependent color frozen (Sec.~\ref{sec:finetune}). Table~\ref{tab:freeze} unfreezes each in turn on three scenes.

\begin{table}[h]
\centering
\small
\caption{\textbf{Freeze ablation of our method}, means over \emph{garden}, \emph{room}, and \emph{bicycle}. Unfreezing the view-dependent color is a tie; unfreezing geometry costs quality even at a tenth of the stock learning rate.}
\label{tab:freeze}
\begin{tabular}{l|c|cccc}
\toprule
also trained besides the atlas & LR scale & PSNR$\uparrow$ & SSIM$\uparrow$ & LPIPS$\downarrow$ & HW FPS$\uparrow$ \\
\midrule
nothing (ours) & --- & 27.70 & 0.8258 & 0.1819 & 3081 \\
SV color & $0.1\times$ & 27.71 & 0.8259 & 0.1819 & 3104 \\
geometry & $0.1\times$ & 27.52 & 0.8134 & 0.1906 & 2908 \\
SV color and geometry & $0.1\times$ & 27.54 & 0.8142 & 0.1899 & 2939 \\
SV color and geometry & $1.0\times$ & 24.99 & 0.6667 & 0.3238 & 2824 \\
\bottomrule
\end{tabular}
\end{table}

\subsection{Contribution of the Density Control}
Our teacher trains with the multi-view criterion of FastGS~\citep{ren2025fastgs} and an absolute-gradient split proxy rather than the positional-gradient rule of 3DGS. Table~\ref{tab:density_control} retrains three scenes with the standard rule and puts them through the identical bake, probe, allocate, and finetune pipeline.

\begin{table}[h]
\centering
\small
\caption{\textbf{Standard density control} in place of the multi-view criterion, all else identical. The criterion is worth $0.67$\,dB and $0.028$ LPIPS on average, and it costs $28\%$ fewer primitives at the same time.}
\label{tab:density_control}
\begin{tabular}{ll|rcccr}
\toprule
scene & density control & $N$ & PSNR$\uparrow$ & SSIM$\uparrow$ & LPIPS$\downarrow$ & SW FPS$\uparrow$ \\
\midrule
garden & ours & 148{,}658 & 27.06 & 0.8405 & 0.1247 & 1933 \\
garden & standard & 240{,}075 & 26.83 & 0.8375 & 0.1303 & 1726 \\
room & ours & 62{,}947 & 31.61 & 0.9216 & 0.1890 & 1966 \\
room & standard & 66{,}547 & 30.32 & 0.9121 & 0.2076 & 2090 \\
bicycle & ours & 160{,}097 & 24.43 & 0.7153 & 0.2321 & 1479 \\
bicycle & standard & 217{,}035 & 23.95 & 0.6648 & 0.2903 & 1547 \\
\midrule
\multicolumn{2}{l|}{mean, ours $-$ standard} & $-28\%$ & $+0.67$ & $+0.0210$ & $-0.0275$ & \\
\bottomrule
\end{tabular}
\end{table}

\subsection{The Per-tile Resolution Cap}
Eq.~\ref{eq:nyquist} clamps each surfel's texture to $r_{\max}=64$ texels per side, and most rectangles reach that cap, which raises the question of whether the cap is what limits quality. Table~\ref{tab:rmax} varies it.

\begin{table}[h]
\centering
\small
\caption{\textbf{Per-tile resolution cap}, means over \emph{garden}, \emph{room}, and \emph{bicycle} through our full pipeline. Doubling the cap doubles the atlas and gains nothing; the lever below $64$ is bytes, not quality.}
\label{tab:rmax}
\begin{tabular}{c|ccc|rrrr}
\toprule
$r_{\max}$ & PSNR$\uparrow$ & SSIM$\uparrow$ & LPIPS$\downarrow$ & $N$ & atlas MB & SW FPS & HW FPS \\
\midrule
32 & 27.71 & 0.8232 & 0.1931 & 123{,}901 & 67 & 1834 & 2922 \\
\textbf{64} (ours) & 27.70 & 0.8258 & \textbf{0.1819} & 123{,}901 & 190 & 1793 & 3081 \\
128 & 27.62 & 0.8209 & 0.1830 & 123{,}901 & 386 & 1747 & 2974 \\
\bottomrule
\end{tabular}
\end{table}

\subsection{Why Textured Gaussians Reads a Flat Frame Rate}
Textured Gaussians~\citep{chao2025textured} reads $28$--$30$ FPS on every scene in Table~\ref{tab:nvs_results}, which looks like a harness artifact. It is in the method: its render call rebuilds the whole texture tensor every frame before the rasterizer sees it, and that cost depends only on the primitive count, which their configuration fixes at $100$K.

\begin{table}[h]
\centering
\small
\caption{\textbf{Where the Textured Gaussians frame goes.} The rebuild is the same $29$\,ms on both scenes, which is why the column is flat. Table~\ref{tab:nvs_results} times the method as its authors wrote it; the cached column is what it could reach with the textures built once, and is not a number we report.}
\label{tab:texgauss}
\begin{tabular}{l|cc|c}
\toprule
scene & full render call & texture rebuild & rasterizer, textures cached \\
\midrule
garden & $33.7$\,ms / $29.7$ FPS & $29.3$\,ms ($87\%$) & $4.2$\,ms / $237$ FPS \\
room & $36.1$\,ms / $27.7$ FPS & $29.3$\,ms ($81\%$) & $6.8$\,ms / $148$ FPS \\
\bottomrule
\end{tabular}
\end{table}

\section{Videos and Demos}
Interactive demos of all 13 benchmark scenes, which render our bakes in the browser and require WebGPU, and split-screen video comparisons against FastGS and Nexels (400K) on \emph{bicycle}, \emph{garden}, and \emph{bonsai}, with a draggable split and zoom, are available on the project page: \url{https://nilkel.github.io/bitymi/}.

\section{Quality--Speed Pareto Fronts}
\label{sec:pareto_supp}
\begin{figure}[t]
  \centering
  \includegraphics[width=\textwidth]{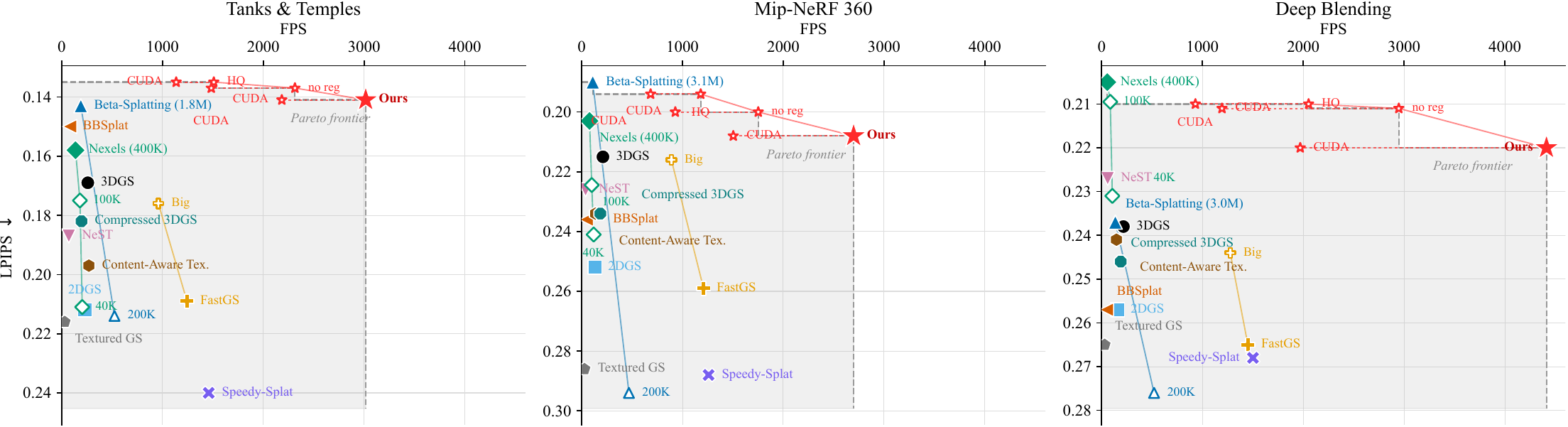}
  \caption{\textbf{LPIPS versus frame rate} for all methods of
  Table~\ref{tab:nvs_results} with their variants. Our points are the
  fixed-function rasterizer; dashed links lead to the same bakes on our CUDA
  renderer. Lower LPIPS is plotted upward, so the best corner is top right.}
  \label{fig:pareto_lpips}
\end{figure}

\begin{figure}[t]
  \centering
  \includegraphics[width=\textwidth]{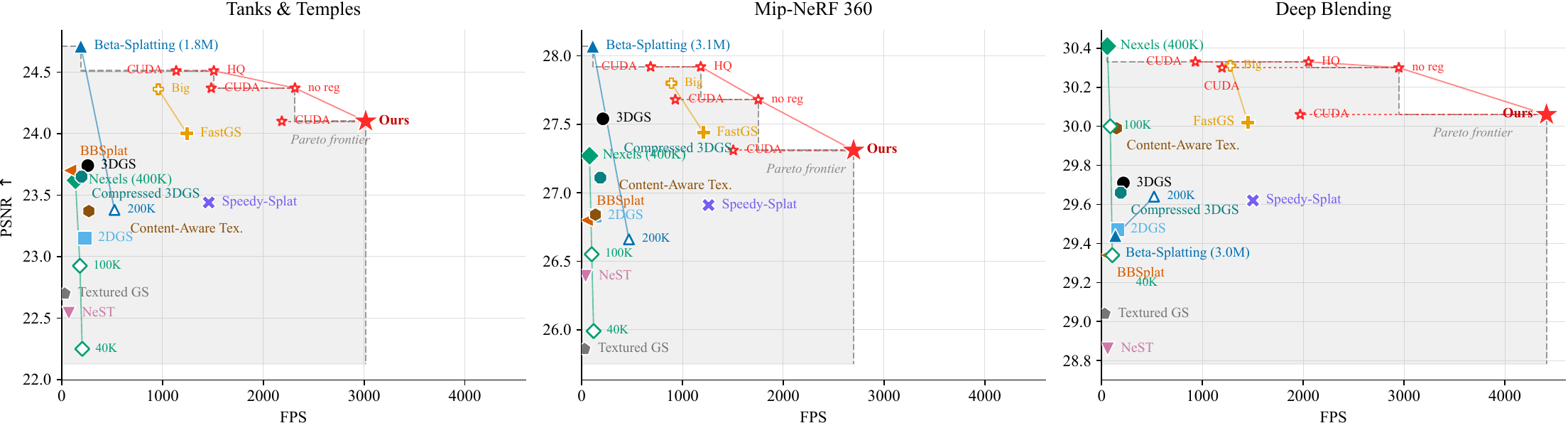}
  \caption{\textbf{PSNR versus frame rate}, same methods and conventions as
  Fig.~\ref{fig:pareto_lpips}.}
  \label{fig:pareto_psnr}
\end{figure}

\begin{figure}[t]
  \centering
  \includegraphics[width=\textwidth]{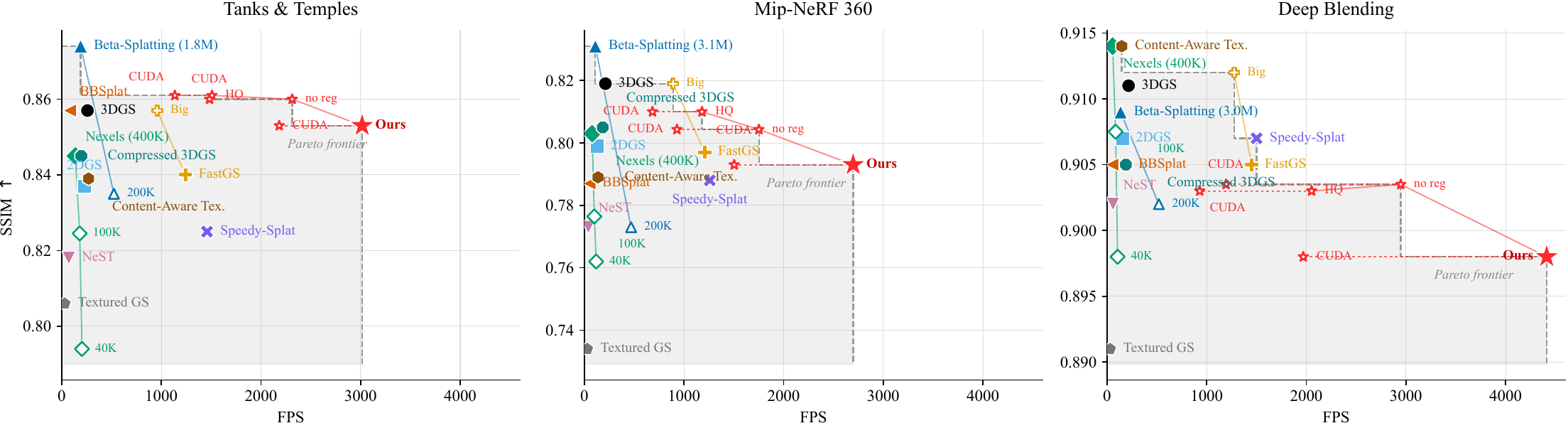}
  \caption{\textbf{SSIM versus frame rate}, same methods and conventions as
  Fig.~\ref{fig:pareto_lpips}.}
  \label{fig:pareto_ssim}
\end{figure}

Figures~\ref{fig:pareto_lpips}--\ref{fig:pareto_ssim} plot every method of
Table~\ref{tab:nvs_results} in the quality--speed plane, one panel per
benchmark, for LPIPS, PSNR, and SSIM against frame rate. Hollow markers are
budget or configuration variants connected to their parent method by thin
segments. Our three configurations are plotted at the fixed-function rates of
Sec.~\ref{sec:hwraster}, the inference path; the dashed links lead to the same
bakes on our CUDA renderer, drawn at the same quality since parity is measured.
Baselines are shown on their authors' own renderers; ported to the same hardware
rasterizer they gain $1.4$--$1.7\times$ (Sec.~\ref{sec:hw_baselines}). The shaded staircase bounds the
Pareto-dominated region. On every benchmark and every metric the high-frame-rate end of the
frontier is ours; on Tanks \& Temples the LPIPS frontier is entirely ours, while
the slow anchors are held by Beta-Splatting, Nexels-400K, or Content-Aware
Texturing depending on the metric.

\section{Per-scene results}

Tables~\ref{tab:ps_psnr}--\ref{tab:ps_size} list per-scene results for every method in the comparison of Table~\ref{tab:nvs_results}, together with FastGS's higher-quality Big configuration and the two variants of Table~\ref{tab:ablation_kernel}, all trained and benchmarked by us on the same RTX 5090 with the shared evaluation script. Table~\ref{tab:falloff_bake_all_scenes} gives the per-scene falloff ablation.

\begin{table}[t]
\centering
\caption{\textbf{Per-scene PSNR$\uparrow$.} Every method trained and benchmarked by us on the same RTX 5090; quality from the shared evaluation script. Best value per scene in bold.}
\label{tab:ps_psnr}
\resizebox{\textwidth}{!}{
\begin{tabular}{l|ccccccccc|cc|cc}
\toprule
& \multicolumn{9}{c|}{\textbf{Mip-NeRF 360}} & \multicolumn{2}{c|}{\textbf{Tanks \& Temples}} & \multicolumn{2}{c}{\textbf{Deep Blending}} \\
\textbf{Method} & bicycle & bonsai & counter & flowers & garden & kitchen & room & stump & treehill & train & truck & drjohnson & playroom \\
\midrule
3DGS & \textbf{25.26} & 32.28 & 29.08 & 21.53 & 27.46 & 31.40 & 31.59 & 26.67 & 22.60 & 22.03 & 25.45 & 29.49 & 29.94 \\
2DGS & 24.72 & 31.29 & 28.11 & 21.17 & 26.70 & 30.34 & 30.68 & 26.17 & 22.30 & 21.20 & 25.10 & 28.87 & 30.06 \\
Beta-Splatting (DBS) & 25.25 & 33.65 & \textbf{30.14} & \textbf{21.75} & \textbf{27.63} & 32.23 & 32.53 & 26.82 & 22.65 & \textbf{22.96} & \textbf{26.46} & 28.65 & 30.24 \\
DBS (200K) & 23.93 & 31.60 & 28.89 & 20.61 & 25.54 & 30.17 & 31.53 & 25.15 & 22.56 & 21.73 & 25.02 & 29.07 & 30.20 \\
Compressed 3DGS & 24.93 & 31.55 & 28.66 & 21.25 & 27.02 & 30.62 & 31.06 & 26.37 & 22.55 & 22.03 & 25.28 & 29.34 & 29.98 \\
\midrule
BBSplat & 23.97 & 31.98 & 28.74 & 20.44 & 26.87 & 31.17 & 31.31 & 24.39 & 22.36 & 22.05 & 25.36 & 28.83 & 29.85 \\
Nexels (40K) & 23.45 & 29.91 & 27.57 & 19.81 & 25.63 & 29.63 & 30.37 & 25.10 & 22.45 & 20.71 & 23.79 & 29.11 & 29.57 \\
Nexels (100K) & 23.88 & 30.88 & 28.20 & 20.26 & 26.02 & 30.51 & 31.07 & 25.52 & 22.62 & 21.11 & 24.74 & 29.76 & 30.24 \\
Nexels (400K) & 24.74 & 31.30 & 28.90 & 21.17 & 26.96 & 31.18 & 31.67 & 26.58 & \textbf{22.93} & 21.70 & 25.54 & \textbf{30.02} & 30.80 \\
Content-Aware Tex. & 23.84 & 31.37 & 28.72 & 20.07 & 26.45 & 30.94 & 31.76 & 25.55 & 22.82 & 21.33 & 25.41 & 29.62 & 30.36 \\
Textured Gaussians & 22.75 & 30.84 & 28.03 & 19.66 & 24.81 & 29.85 & 30.86 & 23.73 & 22.20 & 20.57 & 24.82 & 28.40 & 29.69 \\
NeST-Splatting & 24.36 & 31.91 & 28.32 & 19.42 & 26.59 & 30.29 & 30.98 & 24.52 & 21.09 & 20.54 & 24.55 & 29.27 & 28.44 \\
\midrule
Speedy-Splat & 24.86 & 31.03 & 28.18 & 21.42 & 26.92 & 30.03 & 30.65 & 26.57 & 22.51 & 21.68 & 25.19 & 29.13 & 30.10 \\
FastGS & 24.61 & 32.05 & 29.06 & 21.29 & 27.32 & 31.64 & 31.70 & 26.47 & 22.83 & 22.31 & 25.70 & 29.59 & 30.44 \\
FastGS (Big) & 24.99 & 32.81 & 29.46 & 21.63 & 27.23 & 32.27 & 31.91 & \textbf{27.01} & 22.85 & 22.68 & 26.05 & 29.77 & \textbf{30.85} \\
\midrule
Ours & 24.43 & 32.69 & 29.36 & 20.76 & 27.06 & 31.52 & 31.61 & 25.82 & 22.56 & 22.60 & 25.61 & 29.62 & 30.49 \\
Ours (HQ) & 24.79 & 33.79 & 30.04 & 21.15 & 27.52 & 32.51 & 32.59 & 26.34 & 22.55 & 22.97 & 26.05 & 29.88 & 30.77 \\
Ours (no reg.) & 24.70 & 33.36 & 29.78 & 20.92 & 27.43 & 32.12 & 32.08 & 26.19 & 22.58 & 22.87 & 25.87 & 29.84 & 30.75 \\
\bottomrule
\end{tabular}
}
\end{table}

\begin{table}[t]
\centering
\caption{\textbf{Per-scene SSIM$\uparrow$.} Every method trained and benchmarked by us on the same RTX 5090; quality from the shared evaluation script. Best value per scene in bold.}
\label{tab:ps_ssim}
\resizebox{\textwidth}{!}{
\begin{tabular}{l|ccccccccc|cc|cc}
\toprule
& \multicolumn{9}{c|}{\textbf{Mip-NeRF 360}} & \multicolumn{2}{c|}{\textbf{Tanks \& Temples}} & \multicolumn{2}{c}{\textbf{Deep Blending}} \\
\textbf{Method} & bicycle & bonsai & counter & flowers & garden & kitchen & room & stump & treehill & train & truck & drjohnson & playroom \\
\midrule
3DGS & 0.768 & 0.944 & 0.911 & 0.608 & 0.869 & 0.930 & 0.923 & 0.774 & 0.644 & 0.825 & 0.889 & 0.907 & 0.914 \\
2DGS & 0.733 & 0.933 & 0.894 & 0.579 & 0.844 & 0.919 & 0.910 & 0.757 & 0.620 & 0.795 & 0.878 & 0.903 & 0.912 \\
Beta-Splatting (DBS) & \textbf{0.784} & \textbf{0.954} & \textbf{0.924} & \textbf{0.635} & \textbf{0.871} & \textbf{0.934} & \textbf{0.935} & \textbf{0.796} & \textbf{0.646} & \textbf{0.844} & \textbf{0.903} & 0.899 & \textbf{0.919} \\
DBS (200K) & 0.676 & 0.935 & 0.901 & 0.531 & 0.772 & 0.912 & 0.917 & 0.706 & 0.603 & 0.802 & 0.867 & 0.896 & 0.909 \\
Compressed 3DGS & 0.753 & 0.935 & 0.898 & 0.588 & 0.852 & 0.916 & 0.913 & 0.760 & 0.625 & 0.811 & 0.879 & 0.903 & 0.907 \\
\midrule
BBSplat & 0.689 & 0.949 & 0.914 & 0.533 & 0.839 & 0.932 & 0.932 & 0.687 & 0.613 & 0.830 & 0.884 & 0.898 & 0.912 \\
Nexels (40K) & 0.658 & 0.915 & 0.881 & 0.496 & 0.797 & 0.903 & 0.907 & 0.691 & 0.608 & 0.753 & 0.835 & 0.899 & 0.897 \\
Nexels (100K) & 0.681 & 0.926 & 0.895 & 0.528 & 0.808 & 0.915 & 0.918 & 0.714 & 0.603 & 0.785 & 0.864 & 0.908 & 0.907 \\
Nexels (400K) & 0.732 & 0.924 & 0.906 & 0.580 & 0.843 & 0.923 & 0.924 & 0.765 & 0.630 & 0.807 & 0.882 & \textbf{0.912} & 0.915 \\
Content-Aware Tex. & 0.707 & 0.935 & 0.902 & 0.513 & 0.831 & 0.919 & 0.924 & 0.733 & 0.634 & 0.798 & 0.880 & 0.910 & 0.918 \\
Textured Gaussians & 0.601 & 0.923 & 0.880 & 0.473 & 0.736 & 0.898 & 0.906 & 0.625 & 0.560 & 0.755 & 0.856 & 0.884 & 0.898 \\
NeST-Splatting & 0.728 & 0.927 & 0.886 & 0.510 & 0.843 & 0.916 & 0.908 & 0.676 & 0.565 & 0.773 & 0.863 & 0.903 & 0.902 \\
\midrule
Speedy-Splat & 0.725 & 0.921 & 0.872 & 0.584 & 0.835 & 0.894 & 0.899 & 0.770 & 0.595 & 0.777 & 0.873 & 0.902 & 0.912 \\
FastGS & 0.724 & 0.937 & 0.899 & 0.572 & 0.848 & 0.922 & 0.912 & 0.753 & 0.606 & 0.805 & 0.875 & 0.900 & 0.909 \\
FastGS (Big) & 0.765 & 0.947 & 0.910 & 0.614 & 0.867 & 0.933 & 0.923 & 0.785 & 0.627 & 0.826 & 0.889 & 0.908 & 0.915 \\
\midrule
Ours & 0.715 & 0.940 & 0.904 & 0.558 & 0.841 & 0.921 & 0.922 & 0.729 & 0.605 & 0.826 & 0.880 & 0.895 & 0.901 \\
Ours (HQ) & 0.740 & 0.948 & 0.915 & 0.590 & 0.855 & 0.932 & 0.929 & 0.759 & 0.620 & 0.836 & 0.887 & 0.900 & 0.906 \\
Ours (no reg.) & 0.732 & 0.947 & 0.912 & 0.573 & 0.853 & 0.929 & 0.927 & 0.751 & 0.615 & 0.833 & 0.887 & 0.900 & 0.907 \\
\bottomrule
\end{tabular}
}
\end{table}

\begin{table}[t]
\centering
\caption{\textbf{Per-scene LPIPS$\downarrow$.} Every method trained and benchmarked by us on the same RTX 5090; quality from the shared evaluation script. Best value per scene in bold.}
\label{tab:ps_lpips}
\resizebox{\textwidth}{!}{
\begin{tabular}{l|ccccccccc|cc|cc}
\toprule
& \multicolumn{9}{c|}{\textbf{Mip-NeRF 360}} & \multicolumn{2}{c|}{\textbf{Tanks \& Temples}} & \multicolumn{2}{c}{\textbf{Deep Blending}} \\
\textbf{Method} & bicycle & bonsai & counter & flowers & garden & kitchen & room & stump & treehill & train & truck & drjohnson & playroom \\
\midrule
3DGS & 0.208 & 0.203 & 0.199 & 0.336 & 0.106 & 0.126 & 0.218 & 0.214 & 0.325 & 0.196 & 0.142 & 0.236 & 0.239 \\
2DGS & 0.270 & 0.226 & 0.230 & 0.372 & 0.147 & 0.146 & 0.243 & 0.259 & 0.374 & 0.251 & 0.172 & 0.256 & 0.257 \\
Beta-Splatting (DBS) & \textbf{0.175} & 0.181 & 0.173 & 0.298 & \textbf{0.102} & 0.122 & 0.192 & \textbf{0.184} & 0.285 & 0.180 & \textbf{0.107} & 0.244 & 0.231 \\
DBS (200K) & 0.348 & 0.224 & 0.219 & 0.436 & 0.271 & 0.158 & 0.232 & 0.329 & 0.430 & 0.246 & 0.183 & 0.280 & 0.272 \\
Compressed 3DGS & 0.233 & 0.213 & 0.213 & 0.354 & 0.135 & 0.138 & 0.228 & 0.245 & 0.349 & 0.211 & 0.152 & 0.244 & 0.248 \\
\midrule
BBSplat & 0.282 & 0.168 & 0.174 & 0.389 & 0.137 & 0.117 & \textbf{0.178} & 0.323 & 0.352 & 0.176 & 0.124 & 0.264 & 0.250 \\
Nexels (40K) & 0.288 & 0.200 & 0.206 & 0.365 & 0.163 & 0.147 & 0.209 & 0.283 & 0.310 & 0.258 & 0.165 & 0.234 & 0.227 \\
Nexels (100K) & 0.267 & 0.184 & 0.187 & 0.333 & 0.153 & 0.131 & 0.194 & 0.268 & 0.303 & 0.220 & 0.130 & 0.215 & 0.204 \\
Nexels (400K) & 0.227 & 0.190 & 0.176 & \textbf{0.297} & 0.124 & 0.123 & 0.186 & 0.216 & \textbf{0.285} & 0.201 & 0.114 & \textbf{0.210} & \textbf{0.199} \\
Content-Aware Tex. & 0.252 & 0.209 & 0.209 & 0.378 & 0.145 & 0.143 & 0.210 & 0.258 & 0.300 & 0.245 & 0.150 & 0.247 & 0.235 \\
Textured Gaussians & 0.345 & 0.219 & 0.218 & 0.412 & 0.236 & 0.165 & 0.223 & 0.359 & 0.396 & 0.256 & 0.176 & 0.270 & 0.261 \\
NeST-Splatting & 0.236 & 0.175 & 0.201 & 0.355 & 0.129 & 0.126 & 0.196 & 0.289 & 0.324 & 0.230 & 0.143 & 0.236 & 0.218 \\
\midrule
Speedy-Splat & 0.304 & 0.251 & 0.273 & 0.394 & 0.183 & 0.201 & 0.278 & 0.260 & 0.445 & 0.290 & 0.190 & 0.266 & 0.269 \\
FastGS & 0.287 & 0.214 & 0.220 & 0.388 & 0.157 & 0.138 & 0.240 & 0.272 & 0.414 & 0.242 & 0.177 & 0.270 & 0.261 \\
FastGS (Big) & 0.220 & 0.188 & 0.196 & 0.323 & 0.106 & 0.117 & 0.216 & 0.216 & 0.363 & 0.211 & 0.141 & 0.247 & 0.240 \\
\midrule
Ours & 0.232 & 0.170 & 0.181 & 0.317 & 0.125 & 0.122 & 0.189 & 0.249 & 0.287 & \textbf{0.168} & 0.113 & 0.228 & 0.212 \\
Ours (HQ) & 0.212 & 0.160 & 0.169 & 0.297 & 0.112 & 0.110 & 0.179 & 0.221 & 0.283 & 0.164 & 0.106 & 0.217 & 0.202 \\
Ours (no reg.) & 0.221 & 0.167 & 0.174 & 0.308 & 0.115 & 0.115 & 0.185 & 0.227 & 0.286 & 0.166 & 0.108 & 0.219 & 0.202 \\
\bottomrule
\end{tabular}
}
\end{table}

\begin{table}[t]
\centering
\caption{\textbf{Per-scene FPS$\uparrow$.} Every method trained and benchmarked by us on the same RTX 5090; quality from the shared evaluation script. Best value per scene in bold. Ours is listed twice: SW is the CUDA renderer of Sec.~\ref{sec:conic}, HW the hardware rasterizer of Sec.~\ref{sec:hwraster} on the same bakes. Baselines run on their authors' own renderers (Compressed 3DGS also on its hardware-rasterizing viewer, marked HW).}
\label{tab:ps_fps}
\resizebox{\textwidth}{!}{
\begin{tabular}{l|ccccccccc|cc|cc}
\toprule
& \multicolumn{9}{c|}{\textbf{Mip-NeRF 360}} & \multicolumn{2}{c|}{\textbf{Tanks \& Temples}} & \multicolumn{2}{c}{\textbf{Deep Blending}} \\
\textbf{Method} & bicycle & bonsai & counter & flowers & garden & kitchen & room & stump & treehill & train & truck & drjohnson & playroom \\
\midrule
3DGS & 111 & 357 & 258 & 224 & 145 & 211 & 244 & 167 & 184 & 273 & 244 & 165 & 267 \\
2DGS & 91 & 132 & 181 & 141 & 106 & 153 & 165 & 99 & 108 & 241 & 220 & 154 & 159 \\
Beta-Splatting (DBS) & 58 & 154 & 129 & 102 & 79 & 141 & 156 & 71 & 99 & 254 & 126 & 114 & 158 \\
DBS (200K) & 454 & 478 & 418 & 489 & 569 & 412 & 434 & 490 & 474 & 534 & 513 & 516 & 523 \\
Compressed 3DGS & 143 & 218 & 184 & 208 & 170 & 172 & 184 & 194 & 190 & 194 & 200 & 172 & 206 \\
Compressed 3DGS (HW) & 797 & 1198 & 956 & 1158 & 742 & 792 & 1068 & 1049 & 1092 & 1211 & 1154 & 1035 & 1310 \\
\midrule
BBSplat & 31 & 69 & 66 & 27 & 57 & 74 & 54 & 26 & 30 & 86 & 74 & 66 & 33 \\
Nexels (40K) & 141 & 86 & 96 & 138 & 143 & 98 & 94 & 142 & 128 & 205 & 203 & 107 & 104 \\
Nexels (100K) & 117 & 70 & 82 & 119 & 121 & 86 & 79 & 121 & 106 & 181 & 181 & 87 & 84 \\
Nexels (400K) & 91 & 51 & 66 & 93 & 96 & 74 & 62 & 93 & 78 & 139 & 135 & 59 & 54 \\
Content-Aware Tex. & 156 & 84 & 96 & 156 & 221 & 90 & 133 & 137 & 157 & 297 & 242 & 161 & 133 \\
Textured Gaussians & 30 & 28 & 26 & 28 & 30 & 27 & 28 & 29 & 28 & 30 & 30 & 30 & 30 \\
NeST-Splatting & 45 & 38 & 37 & 34 & 44 & 43 & 42 & 37 & 33 & 75 & 67 & 52 & 68 \\
\midrule
Speedy-Splat & 908 & 1478 & 1413 & 1196 & 1006 & 1414 & 1479 & 1062 & 1369 & 1508 & 1410 & 1390 & 1612 \\
FastGS & 1212 & 1229 & 1199 & 1166 & 1146 & 1055 & 1340 & 1263 & 1266 & 1139 & 1348 & 1403 & 1498 \\
FastGS (Big) & 837 & 898 & 1031 & 873 & 584 & 700 & 1116 & 970 & 997 & 928 & 991 & 1232 & 1320 \\
\midrule
Ours (SW) & 1479 & 1288 & 1562 & 1263 & 1933 & 1354 & 1966 & 1419 & 1270 & 1996 & 2392 & 2109 & 1827 \\
\midrule
\textbf{Ours (HW)} & \textbf{2988} & \textbf{2650} & \textbf{2294} & \textbf{2313} & \textbf{2842} & \textbf{1771} & \textbf{3413} & \textbf{3406} & \textbf{2607} & \textbf{2769} & \textbf{3262} & \textbf{4323} & \textbf{4503} \\
Ours (HQ, HW) & 1356 & 1134 & 1032 & 1046 & 1308 & 821 & 1324 & 1408 & 1198 & 1436 & 1582 & 2051 & 2055 \\
Ours (no reg., HW) & 1990 & 1774 & 1530 & 1585 & 1868 & 1173 & 2190 & 2037 & 1621 & 2239 & 2389 & 3038 & 2858 \\
\bottomrule
\end{tabular}
}
\end{table}

\begin{table}[t]
\centering
\caption{\textbf{Per-scene primitive counts.} Every method trained and benchmarked by us on the same RTX 5090; quality from the shared evaluation script. Reported for context, not ranked.}
\label{tab:ps_prims}
\resizebox{\textwidth}{!}{
\begin{tabular}{l|ccccccccc|cc|cc}
\toprule
& \multicolumn{9}{c|}{\textbf{Mip-NeRF 360}} & \multicolumn{2}{c|}{\textbf{Tanks \& Temples}} & \multicolumn{2}{c}{\textbf{Deep Blending}} \\
\textbf{Method} & bicycle & bonsai & counter & flowers & garden & kitchen & room & stump & treehill & train & truck & drjohnson & playroom \\
\midrule
3DGS & 4.87M & 1.08M & 1.08M & 2.91M & 4.14M & 1.61M & 1.31M & 4.28M & 3.30M & 1.09M & 2.06M & 3.10M & 1.84M \\
2DGS & 4.57M & 0.80M & 0.65M & 2.26M & 3.00M & 0.83M & 0.86M & 2.96M & 3.23M & 0.56M & 1.14M & 1.81M & 1.21M \\
Beta-Splatting (DBS) & 6.00M & 1.50M & 1.50M & 3.00M & 5.00M & 1.50M & 1.50M & 4.50M & 3.50M & 1.00M & 2.50M & 3.50M & 2.50M \\
DBS (200K) & 0.20M & 0.21M & 0.20M & 0.20M & 0.20M & 0.24M & 0.20M & 0.20M & 0.20M & 0.20M & 0.20M & 0.20M & 0.20M \\
Compressed 3DGS & 4.72M & 1.04M & 1.05M & 2.87M & 4.11M & 1.57M & 1.24M & 4.18M & 3.23M & 1.06M & 2.00M & 2.97M & 1.80M \\
\midrule
BBSplat & 0.30M & 0.16M & 0.16M & 0.30M & 0.30M & 0.16M & 0.16M & 0.30M & 0.30M & 0.30M & 0.30M & 0.16M & 0.16M \\
Nexels (40K) & 0.04M & 0.04M & 0.04M & 0.04M & 0.04M & 0.04M & 0.04M & 0.04M & 0.04M & 0.04M & 0.04M & 0.04M & 0.04M \\
Nexels (100K) & 0.10M & 0.10M & 0.10M & 0.10M & 0.10M & 0.10M & 0.10M & 0.10M & 0.10M & 0.10M & 0.10M & 0.10M & 0.10M \\
Nexels (400K) & 0.40M & 0.40M & 0.40M & 0.40M & 0.40M & 0.40M & 0.40M & 0.40M & 0.40M & 0.40M & 0.40M & 0.40M & 0.40M \\
Content-Aware Tex. & 0.11M & 0.23M & 0.20M & 0.10M & 0.12M & 0.25M & 0.17M & 0.13M & 0.10M & 0.15M & 0.12M & 0.23M & 0.14M \\
Textured Gaussians & 0.10M & 0.10M & 0.10M & 0.10M & 0.10M & 0.10M & 0.10M & 0.10M & 0.10M & 0.10M & 0.10M & 0.10M & 0.10M \\
NeST-Splatting & 2.09M & 0.34M & 0.28M & 1.44M & 1.52M & 0.43M & 0.34M & 0.80M & 1.42M & 0.26M & 0.50M & 0.62M & 0.34M \\
\midrule
Speedy-Splat & 0.61M & 0.13M & 0.10M & 0.36M & 0.53M & 0.12M & 0.12M & 0.50M & 0.37M & 0.11M & 0.26M & 0.31M & 0.19M \\
FastGS & 0.54M & 0.28M & 0.21M & 0.49M & 0.66M & 0.38M & 0.21M & 0.39M & 0.39M & 0.23M & 0.25M & 0.25M & 0.18M \\
FastGS (Big) & 1.56M & 0.85M & 0.47M & 1.13M & 2.63M & 1.18M & 0.58M & 1.05M & 1.00M & 0.46M & 0.63M & 0.71M & 0.59M \\
\midrule
Ours & 0.16M & 0.09M & 0.07M & 0.18M & 0.15M & 0.12M & 0.06M & 0.10M & 0.17M & 0.09M & 0.09M & 0.07M & 0.07M \\
Ours (HQ) & 0.34M & 0.17M & 0.13M & 0.38M & 0.29M & 0.24M & 0.13M & 0.18M & 0.37M & 0.15M & 0.15M & 0.14M & 0.12M \\
Ours (no reg.) & 0.21M & 0.13M & 0.09M & 0.24M & 0.21M & 0.17M & 0.09M & 0.13M & 0.24M & 0.10M & 0.11M & 0.09M & 0.09M \\
\bottomrule
\end{tabular}
}
\end{table}

\begin{table}[t]
\centering
\caption{\textbf{Per-scene stored render payload (MB).} Every method trained and benchmarked by us on the same RTX 5090; quality from the shared evaluation script. Reported for context, not ranked; formats are mixed (see Table~\ref{tab:nvs_results}). The last row gives the sized BC7 atlas portion of our payload alone.}
\label{tab:ps_size}
\resizebox{\textwidth}{!}{
\begin{tabular}{l|ccccccccc|cc|cc}
\toprule
& \multicolumn{9}{c|}{\textbf{Mip-NeRF 360}} & \multicolumn{2}{c|}{\textbf{Tanks \& Temples}} & \multicolumn{2}{c}{\textbf{Deep Blending}} \\
\textbf{Method} & bicycle & bonsai & counter & flowers & garden & kitchen & room & stump & treehill & train & truck & drjohnson & playroom \\
\midrule
3DGS & 1152 & 255 & 255 & 689 & 980 & 382 & 311 & 1012 & 781 & 258 & 486 & 734 & 435 \\
2DGS & 1064 & 186 & 150 & 525 & 699 & 192 & 200 & 690 & 751 & 130 & 266 & 421 & 281 \\
Beta-Splatting (DBS) & 687 & 172 & 172 & 343 & 572 & 172 & 172 & 515 & 401 & 114 & 286 & 401 & 286 \\
DBS (200K) & 23 & 24 & 23 & 23 & 23 & 28 & 23 & 23 & 23 & 23 & 23 & 23 & 23 \\
Compressed 3DGS & 47 & 13 & 14 & 31 & 41 & 19 & 15 & 41 & 35 & 16 & 22 & 31 & 19 \\
\midrule
BBSplat & 215 & 124 & 124 & 209 & 245 & 130 & 118 & 206 & 212 & 227 & 224 & 119 & 104 \\
Nexels (40K) & 138 & 138 & 138 & 138 & 138 & 138 & 138 & 138 & 138 & 138 & 138 & 138 & 138 \\
Nexels (100K) & 282 & 282 & 282 & 282 & 282 & 282 & 282 & 282 & 282 & 282 & 282 & 282 & 282 \\
Nexels (400K) & 230 & 230 & 230 & 230 & 230 & 230 & 230 & 230 & 230 & 230 & 230 & 230 & 230 \\
Content-Aware Tex. & 703 & 145 & 167 & 593 & 456 & 156 & 204 & 877 & 625 & 106 & 148 & 255 & 248 \\
Textured Gaussians & 3837 & 3837 & 3837 & 3837 & 3837 & 3837 & 3837 & 3837 & 3837 & 3837 & 3837 & 3837 & 3837 \\
NeST-Splatting & 487 & 78 & 65 & 336 & 353 & 99 & 79 & 186 & 330 & 61 & 118 & 146 & 80 \\
\midrule
Speedy-Splat & 145 & 31 & 23 & 86 & 126 & 28 & 28 & 119 & 87 & 25 & 61 & 74 & 44 \\
FastGS & 134 & 68 & 52 & 122 & 164 & 94 & 51 & 97 & 98 & 57 & 63 & 63 & 45 \\
FastGS (Big) & 388 & 210 & 117 & 281 & 652 & 293 & 143 & 261 & 249 & 114 & 155 & 177 & 146 \\
\midrule
Ours & 324 & 118 & 95 & 402 & 323 & 125 & 89 & 202 & 376 & 172 & 170 & 122 & 85 \\
Ours (HQ) & 698 & 270 & 206 & 923 & 725 & 298 & 255 & 387 & 686 & 321 & 373 & 306 & 206 \\
Ours (no reg.) & 488 & 201 & 171 & 641 & 553 & 207 & 171 & 297 & 593 & 236 & 268 & 222 & 155 \\
\midrule
Ours (atlas only) & 252 & 78 & 65 & 320 & 257 & 71 & 61 & 159 & 298 & 132 & 132 & 92 & 55 \\
\bottomrule
\end{tabular}
}
\end{table}

\begin{table}[t]
\centering
\caption{\textbf{Falloff regularizer, per scene.} Mip-NeRF 360, full pipeline (trained, baked, texture-finetuned) on a fixed allocation, plain prior at each $\lambda$. Ablation arms use the previous shelf allocation, so absolute values differ from Table~\ref{tab:nvs_results}; the comparison is within each pair. $N$ is the primitive count and HW the hardware frame rate on an RTX 5090.}
\label{tab:falloff_bake_all_scenes}
\resizebox{\textwidth}{!}{
\begin{tabular}{l|rccr|rccr|rccr|rccr}
\toprule
& \multicolumn{4}{c|}{$\lambda = 0$} & \multicolumn{4}{c|}{$\lambda = 0.005$ \emph{(ours)}} & \multicolumn{4}{c|}{$\lambda = 0.01$} & \multicolumn{4}{c}{$\lambda = 0.05$} \\
Scene & $N$ & PSNR & LPIPS & HW & $N$ & PSNR & LPIPS & HW & $N$ & PSNR & LPIPS & HW & $N$ & PSNR & LPIPS & HW \\
\midrule
bicycle & 216{,}452 & 24.65 & 0.2277 & 2009 & 156{,}646 & 24.40 & 0.2407 & 2992 & 142{,}403 & 24.18 & 0.2523 & 3424 & 150{,}834 & 22.60 & 0.3280 & 4258 \\
bonsai & 130{,}163 & 33.39 & 0.1693 & 1682 & 92{,}104 & 32.67 & 0.1701 & 2503 & 82{,}226 & 32.04 & 0.1783 & 3059 & 88{,}835 & 28.20 & 0.2493 & 3782 \\
counter & 92{,}705 & 29.72 & 0.1783 & 1508 & 68{,}018 & 29.30 & 0.1843 & 2212 & 59{,}906 & 28.87 & 0.1946 & 2740 & 69{,}802 & 26.52 & 0.2600 & 3426 \\
flowers & 238{,}688 & 20.95 & 0.3134 & 1600 & 183{,}101 & 20.76 & 0.3213 & 2334 & 175{,}074 & 20.59 & 0.3314 & 2726 & 196{,}840 & 19.42 & 0.3928 & 3108 \\
garden & 210{,}985 & 27.37 & 0.1194 & 1845 & 148{,}723 & 27.03 & 0.1294 & 2770 & 137{,}102 & 26.72 & 0.1384 & 3232 & 160{,}818 & 24.79 & 0.2046 & 3387 \\
kitchen & 171{,}587 & 32.13 & 0.1163 & 1144 & 118{,}917 & 31.46 & 0.1236 & 1726 & 99{,}556 & 30.83 & 0.1345 & 2072 & 99{,}170 & 28.47 & 0.1933 & 2778 \\
room & 90{,}200 & 32.16 & 0.1869 & 2164 & 62{,}388 & 31.51 & 0.1898 & 3337 & 58{,}134 & 31.00 & 0.1984 & 3815 & 69{,}156 & 28.51 & 0.2599 & 4898 \\
stump & 125{,}208 & 26.09 & 0.2406 & 2027 & 93{,}842 & 25.79 & 0.2621 & 3196 & 94{,}960 & 25.40 & 0.2826 & 3858 & 120{,}502 & 23.51 & 0.3864 & 4057 \\
treehill & 237{,}410 & 22.59 & 0.2932 & 1605 & 169{,}265 & 22.44 & 0.2969 & 2705 & 156{,}047 & 22.44 & 0.3038 & 3153 & 160{,}750 & 21.32 & 0.3542 & 4001 \\
\midrule
mean & 168{,}155 & 27.67 & 0.2050 & 1732 & 121{,}445 & 27.26 & 0.2131 & 2642 & 111{,}712 & 26.90 & 0.2238 & 3123 & 124{,}079 & 24.82 & 0.2921 & 3744 \\
\bottomrule
\end{tabular}}
\end{table}

\end{document}